\RequirePackage{letltxmacro}
\documentclass[pdflatex,sn-mathphys]{sn-jnl}

\jyear{2021}%

\theoremstyle{thmstyleone}%
\usepackage[T5]{fontenc}
\usepackage[utf8]{inputenc}

\usepackage{booktabs}
\usepackage{float}
\usepackage{adjustbox}
\usepackage{caption}
\usepackage{subcaption}
\usepackage{graphicx}
\usepackage{amsmath}
\usepackage{algorithm}
\usepackage{xcolor}
\usepackage{algpseudocode}
\usepackage{threeparttable}
\usepackage{array} 

\usepackage{xcolor}
\usepackage{framed}
\usepackage{enumitem}

\definecolor{prompttitle}{gray}{0.0}

\newenvironment{promptcard}[1]{%
    \par\medskip
    \MakeFramed{\advance\hsize-\width \FrameRestore}%
    \noindent
    \colorbox{black}{%
        \parbox{\dimexpr\linewidth-2\fboxsep\relax}{%
            \color{white}\bfseries\small #1%
        }%
    }%
    \par\smallskip
    \small
}{%
    \endMakeFramed
    \par\medskip
}

\newcommand{\promptsection}[1]{%
    \par\smallskip\noindent\textbf{#1.}\quad%
}

\usepackage{amssymb}
\usepackage{pifont}
\newcommand{\cmark}{\ding{51}}%
\newcommand{\xmark}{\ding{55}}%

\theoremstyle{thmstyletwo}%

\theoremstyle{thmstylethree}%

\definecolor{lavender}{RGB}{240,230,250}
\definecolor{lightblue}{RGB}{173, 216, 230}

\begin{document}


\title[Bao et al.]{OpenViVE: New Benchmark and Dual-Stream Co-Attention Framework for Vietnamese Visual Entailment}


\author[1,2]{\fnm{Bao} \sur{Quoc Trinh}}\email{22520125@gm.uit.edu.vn}

\author*[1,2]{\fnm{Tin} \sur{Van Huynh}}\email{tinhv@uit.edu.vn}

\author[1,2]{\fnm{Kiet} \sur{Van Nguyen}}\email{kietnv@uit.edu.vn}

\author[1,2]{\fnm{Ngan} \sur{Luu-Thuy Nguyen}}\email{ngannlt@uit.edu.vn}

\affil[1]{\orgname{University of Information Technology, Ho Chi Minh City, Vietnam}}
\affil[2]{\orgname{Vietnam National University, Ho Chi Minh City, Vietnam}}


\abstract{Visual Entailment (VE) determines whether a textual hypothesis is entailed, contradicted, or unverifiable from an image. However, existing VE benchmarks are largely English-centric and do not capture Vietnamese-specific linguistic distinctions or culturally grounded visual contexts. To address this gap, we introduce OpenViVE, the first large-scale native Vietnamese VE benchmark, containing 62,268 balanced image–hypothesis pairs from 10,378 culturally grounded images. OpenViVE is constructed using an Artifact-Aware Hybrid LLM–Human Pipeline that combines constrained generation, human validation, and PMI-guided refinement. The pipeline substantially reduces hypothesis-only shortcuts, with Contradiction accuracy dropping from 96\% to 66\% after artifact refinement. We further evaluate a dual-stream PhoBERT–ConvNeXt framework with bidirectional co-attention to align visual evidence with Vietnamese hypotheses. The framework achieves 84.84\% accuracy and 84.90\% macro-F1 on the test set, while PhoBERT consistently outperforms multilingual XLM-R variants. These results highlight the importance of language-specific encoding and explicit cross-modal interaction for Vietnamese VE.
}


\keywords{Visual Entailment, Vietnamese NLP, Multimodal reasoning, Low-resource benchmark, Cross-modal attention, Fact verification}

\maketitle

\section{Introduction}
\label{introduction}

The rapid proliferation of digital multimedia content has made semantic verification an increasingly important capability for multimodal information systems. In applications ranging from multimedia fact-checking to cross-modal retrieval, a system must determine not merely whether an image and a piece of text are semantically related, but whether the textual claim is supported, contradicted, or unverifiable given the visual evidence. Visual Entailment (VE) formalizes this problem as a three-way multimodal inference task: given an image premise $I$ and a natural-language hypothesis $H$, a model predicts the semantic relation $y \in \{\text{Entailment}, \text{Contradiction}, \text{Neutral}\}$~\cite{xie2019visual}. Unlike global image--text matching, VE requires fine-grained verification of the propositional content expressed in the hypothesis.

\begin{figure}[!htbp]
\centering
\includegraphics[width=0.85\linewidth]{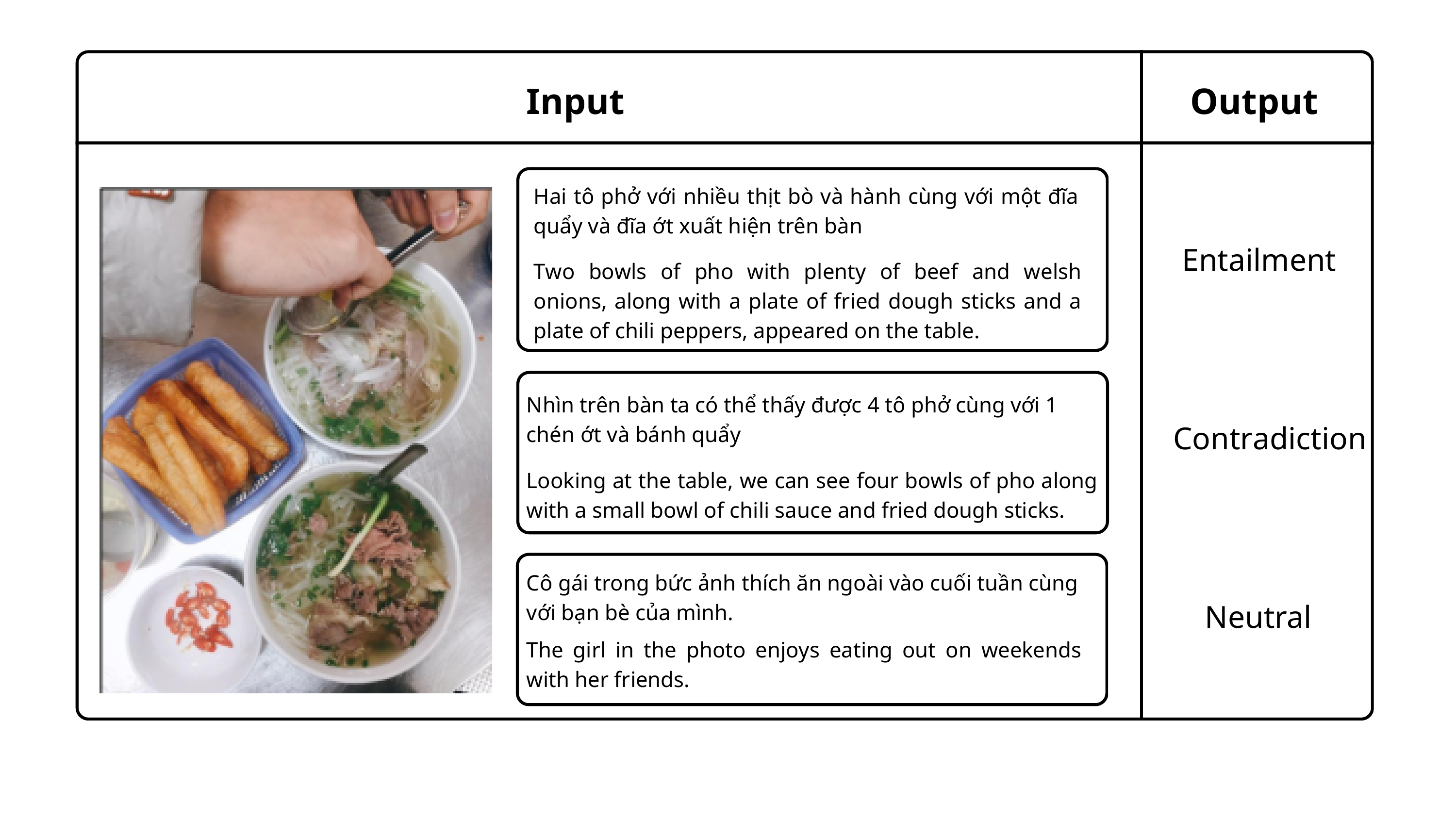}
\caption{A representative example from OpenViVE illustrating the three-way Vietnamese Visual Entailment setting. Given the same premise image, the hypotheses correspond to Entailment, Contradiction, and Neutral according to whether the stated information is supported, contradicted, or unverifiable from the visual evidence. English translations are provided for readability only.}
\label{fig:example_data}
\end{figure}

Existing VE benchmarks, most notably SNLI-VE~\cite{xie2019visual} and e-SNLI-VE~\cite{do2020snli}, have established strong foundations for English visual-textual inference. However, directly extending these resources to Vietnamese through machine translation is insufficient, as Vietnamese visual descriptions often involve language- and culture-specific expressions whose meanings must be grounded in the image. Figure~\ref{fig:example_data} illustrates this challenge through a representative sample from OpenViVE. Given the same premise image, the Entailment hypothesis describes visual evidence that is directly supported by the image, including the visible bowls of \textit{phở} and accompanying food; the Contradiction hypothesis preserves a similar scene description but changes a visually verifiable detail by stating that four bowls of \textit{phở} are present; and the Neutral hypothesis introduces a plausible statement about the person's preference for eating out on weekends, which cannot be verified or refuted from the image alone. The example further highlights language-specific visual grounding in Vietnamese, where terms such as \textit{tô} and \textit{chén} distinguish different container types that may be rendered more generically as ``bowl'' in English. Together with culturally specific entities and scenes such as \textit{quẩy}, \textit{bánh mì}, street-food settings, and Vietnamese streetscapes, these distinctions motivate the construction of a natively Vietnamese VE benchmark rather than relying solely on translated English resources.

Vietnamese vision--language research has expanded across image captioning~\cite{lam2020uitviic,bui2023uitopenviic,nguyen2026linguistically}, visual question answering~\cite{nguyen2023openvivqa,nguyen2024vitextvqa}, and multimodal sentiment analysis~\cite{nguyen2025new}. In parallel, Vietnamese textual Natural Language Inference (NLI) has been studied through benchmarks such as ViNLI~\cite{huynh-etal-2022-vinli}, ViANLI~\cite{van2026new}, and ViLegalNLI~\cite{duong2026vilegalnli}. Nevertheless, these resources do not directly evaluate the three-way semantic relation between a visual premise and a Vietnamese textual hypothesis. Unlike captioning or question answering, VE requires a model to verify the full propositional content of a hypothesis against visual evidence, including attributes, quantities, object identities, actions, and spatial relations. To the best of our knowledge, no large-scale Vietnamese benchmark currently targets three-way Visual Entailment with visually grounded hypotheses, leaving an important gap in the evaluation of Vietnamese multimodal reasoning.

Constructing and modeling such a benchmark introduces two closely related challenges. On the data side, large-scale manual annotation of image--hypothesis pairs is costly because each hypothesis must be jointly evaluated against both the image and its intended semantic label. Although large language models can reduce annotation effort, automatically generated hypotheses may contain hallucinated details, repetitive generation patterns, or label-specific lexical artifacts. Such artifacts are particularly problematic in inference tasks because models may exploit textual shortcuts and predict labels without sufficiently grounding the hypothesis in the image~\cite{gururangan2018annotation}. On the modeling side, Vietnamese VE requires fine-grained interaction between visual evidence and Vietnamese textual representations. Language-specific lexical distinctions and culturally grounded visual content therefore motivate the investigation of Vietnamese-specific text encoders together with explicit cross-modal interaction, rather than relying exclusively on independently encoded or generic multilingual representations.

To address these challenges, we make four main contributions:

\begin{itemize}

    \item \textbf{OpenViVE: a large-scale native Vietnamese VE benchmark.}
We introduce OpenViVE, comprising 62,268 image--hypothesis pairs constructed from 10,378 culturally grounded images. To the best of our knowledge, OpenViVE is the first large-scale benchmark specifically designed for Vietnamese Visual Entailment. Rather than translating hypotheses from existing English VE resources, premise images are drawn from UIT-OpenViIC~\cite{bui2023uitopenviic}, while hypotheses are generated directly in Vietnamese under label-specific constraints designed to preserve Vietnamese lexical distinctions, cultural references, and linguistic naturalness. The dataset is fully balanced across Entailment, Contradiction, and Neutral and is partitioned at the image level to prevent visual leakage across training, development, and test splits.

\item \textbf{Artifact-aware hybrid construction pipeline.}
We propose a semi-automatic data construction pipeline that combines rule-constrained LLM generation using Gemini 2.5 Flash, iterative human validation, and Pointwise Mutual Information (PMI)-guided lexical artifact refinement. Hypothesis-only analysis shows that the refinement process substantially weakens label-predictive textual shortcuts, with Contradiction accuracy decreasing from 0.96 to 0.66 and Entailment accuracy decreasing from 0.96 to 0.80 after artifact reduction. Inter-annotator agreement further reaches Fleiss' $\kappa=0.92$ on the test split, supporting the semantic reliability of the final evaluation data.

\item \textbf{Dual-stream co-attention framework for Vietnamese VE.}
We develop a PhoBERT--ConvNeXt framework that combines spatial visual representations from a frozen ConvNeXt backbone with contextualized Vietnamese textual representations from a fine-tuned PhoBERT encoder through a single bidirectional co-attention fusion block. The framework is designed to explicitly model image--text interaction before classification. Across both Swin-v2 and ConvNeXt visual backbones, PhoBERT-based configurations consistently outperform the corresponding XLM-R variants, providing empirical evidence that Vietnamese-specific textual representations are beneficial for this task. Ablation results further show a substantial performance drop when the co-attention module is removed, highlighting the importance of explicit cross-modal interaction.

\item \textbf{Comprehensive empirical evaluation.}
We benchmark OpenViVE using fine-tuned vision--language baselines, multiple encoder and fusion configurations, zero-shot multimodal LLMs, and fine-tuned multimodal LLMs. The proposed ConvNeXt + PhoBERT$_{\text{large}}$ configuration achieves 84.84\% Accuracy and 84.90\% macro F1-score on the test set, slightly outperforming fine-tuned Qwen3.5-VL-0.8B while substantially surpassing the evaluated zero-shot multimodal LLMs. Ablation studies and behavioral analyses further show that fine-grained verification, particularly the distinction between Entailment and Contradiction, remains a major challenge for Vietnamese Visual Entailment.

\end{itemize}

The remainder of this paper is organized as follows. Section~\ref{sec:related} reviews prior work on Visual Entailment, vision--language evaluation, Vietnamese multimodal resources, and LLM-assisted data construction. Section~\ref{sec:dataset_creation} presents the OpenViVE construction pipeline, validation procedure, and dataset analysis. Section~\ref{sec:framework} describes the PhoBERT--ConvNeXt co-attention framework. Section~\ref{sec:experiments} presents the experimental setup and comparisons with vision--language baselines, encoder variants, and both zero-shot and fine-tuned multimodal LLMs, followed by ablation and behavioral analyses. Finally, Section~\ref{sec:conclusions} concludes the paper and discusses directions for future work.

\section{Related Work}
\label{sec:related}
 
The task of Vietnamese Visual Entailment sits at the intersection of three bodies of
literature: the evaluation methodology of vision--language benchmarks
(Section~\ref{subsec:ve_benchmarks}), the growing ecosystem of Vietnamese multimodal
resources (Section~\ref{subsec:vietnamese}), and the emerging practice of LLM-assisted
multimodal data construction (Section~\ref{subsec:llm_construction}).
We survey each in turn, identifying the specific limitations that OpenViVE is
designed to address and the design principles that follow from each gap.
Table \ref{tab:viet_comparison}
and Table~\ref{tab:construction_comparison} provide structured comparisons of the most
relevant datasets and construction approaches to support precise positioning of
OpenViVE within each literature.
 
\subsection{Visual Entailment Benchmarks and Vision--Language Evaluation}
\label{subsec:ve_benchmarks}
\subsubsection{Visual Entailment Datasets}

Visual Entailment (VE) was introduced by Xie~et~al.~\cite{xie2019visual}
through SNLI-VE, which combines the Flickr30k image collection with
hypotheses derived from the Stanford Natural Language Inference (SNLI)
corpus by replacing textual premises with their corresponding images.
SNLI-VE contains 565,286 image--hypothesis pairs and establishes the
three-way Entailment, Contradiction, and Neutral formulation widely used
in subsequent VE research. However, because the hypotheses were originally
written with respect to textual captions rather than the complete visual
scene, replacing a caption with its corresponding image can change the
intended semantic relation and introduce label inconsistencies.
Do~et~al.~\cite{do2020snli} subsequently addressed this issue by
relabeling problematic examples in SNLI-VE and introducing SNLI-VE-2.0
and e-SNLI-VE, the latter additionally providing human-written natural
language explanations. These studies highlight the importance of validating
image--hypothesis relations directly against visual evidence.

Recent work has extended VE beyond the original SNLI-VE setting.
Defeasible Visual Entailment~\cite{zhang2025defeasible} introduces
additional textual updates that can strengthen or weaken an existing
image--hypothesis relation, requiring models to reason about changes in
entailment under new information. Reijtenbach~et~al.~
\cite{reijtenbach2025dataset} explore a different direction by using
Stable Diffusion to generate synthetic visual premises from SNLI textual
premises. They show that models trained on the synthetic data achieve
performance close to models trained on real images, demonstrating the
potential of generative image models for reducing the cost of VE data
construction. In contrast, OpenViVE retains real-world premise images and
focuses on generating and validating native Vietnamese hypotheses.

Multilingual visual entailment has also begun to receive attention.
IGLUE~\cite{bugliarello2022iglue} introduces XVNLI as a multilingual
grounded entailment benchmark covering English, Arabic, French, Russian,
and Spanish. More recently, Iokawa~et~al.~
\cite{iokawa2024multilingualvte} propose SICK-VTE and QN-VTE to evaluate
multilingual visual-textual entailment under linguistic phenomena such as
quantity and negation. SICK-VTE covers English, Japanese, and Dutch,
whereas QN-VTE focuses on English and Japanese. These resources establish
the importance of evaluating visual entailment beyond English, but they do
not target Vietnamese or large-scale native three-way VE construction.
OpenViVE complements this line of work by providing a large-scale
Vietnamese benchmark whose hypotheses are constructed directly in
Vietnamese over culturally grounded real-world images.
 
\subsubsection{Broader Vision--Language Evaluation}
 
Beyond VE specifically, the broader vision--language evaluation landscape has developed increasingly stringent standards for what constitutes genuine multimodal reasoning, and these standards directly inform the design decisions in OpenViVE.

NLVR2~\cite{suhr2019nlvr2} evaluates structured visual reasoning over image pairs, requiring models to determine whether a statement is true with respect to both images simultaneously; unlike single image VE, it demands relational reasoning across visual contexts. VALSE~\cite{parcalabescu2022valse} provides a task independent benchmark that probes fine grained linguistic phenomena, including existence, plurality, actions, spatial relations, and coreference, against visual grounding. This reveals that many vision language models exhibit near chance performance on specific linguistic categories even when achieving strong aggregate scores elsewhere.
 
The problem of shortcut learning in vision language benchmarks has received increasing attention. SugarCrepe~\cite{hsieh2023sugarcrepe} demonstrates that several compositionality benchmarks contain textual biases that allow hypothesis only models to perform competitively without attending to the image. This vulnerability is directly analogous to the annotation artifact problem in textual NLI. MMStar~\cite{chen2024mmstar} introduces vision indispensability as a dataset design requirement: a sample is vision indispensable only if the correct label cannot be inferred from the hypothesis text alone. Samples that fail this criterion inflate aggregate performance estimates for hypothesis only models and must be identified and removed. MMBench~\cite{liu2023mmbench} further argues that aggregate accuracy obscures significant variation across reasoning types, motivating fine grained capability decomposition. HallusionBench~\cite{guan2024hallusionbench} demonstrates that state of the art large vision language models remain susceptible to language hallucination and visual illusion, motivating benchmarks that expose specific failure modes rather than simply reporting aggregate scores.
 
\subsubsection{Vision--Language Fusion Architectures}

Vision--language models have evolved from joint multimodal encoders toward
modality-specific architectures with explicit cross-modal interaction and,
more recently, large generative multimodal models. Early approaches such as
VisualBERT~\cite{li2019visualbert} and UNITER~\cite{chen2020uniter} jointly
model visual and textual representations within transformer-based multimodal
encoders. ALBEF~\cite{li2021albef} instead separates unimodal representation
learning from multimodal fusion, aligning image and text representations
through contrastive learning before applying cross-modal attention.
TCL~\cite{yang2022tcl} further combines cross-modal alignment with
intra-modal contrastive objectives and local--global representation learning.
These encoder-based approaches provide representative supervised
vision--language baselines for evaluating OpenViVE.

Of particular relevance to our framework, METER~\cite{dou2022meter}
systematically investigates end-to-end vision--language transformer design
using modality-specific visual and textual encoders followed by multimodal
fusion. Its study compares merged-attention and co-attention fusion
strategies and shows that explicit interaction between modality-specific
representations is an effective design for downstream vision--language
tasks. Our framework adopts the general METER-style bidirectional
co-attention principle, but specializes it for Vietnamese Visual Entailment
by combining ConvNeXt visual representations with a Vietnamese-specific
PhoBERT text encoder. Our architectural contribution centers on a specialized dual-stream co-attention design engineered for asymmetric cross-modal alignment between granular spatial visual tokens and compound-word Vietnamese contextual embeddings, empirically validated through systematic modality-interaction ablations.

A complementary direction is represented by multimodal large language
models. BLIP-2~\cite{li2023blip2} connects frozen image encoders and large
language models through a lightweight querying transformer, while
LLaVA~\cite{liu2023llava} demonstrates the effectiveness of multimodal
instruction tuning for general-purpose visual interaction. The Qwen-VL
family~\cite{bai2023qwenvl,bai2025qwen25vl} further extends multimodal
reasoning toward multilingual, text-rich, and fine-grained visual
understanding, with Qwen3-VL providing a more recent generation of the
family for multimodal reasoning.

These developments motivate two complementary evaluation settings in
OpenViVE: specialized encoder-based architectures that explicitly model
image--text interaction, and general-purpose multimodal LLMs evaluated
under both zero-shot and task-specific adaptation. Their relative
effectiveness for Vietnamese three-way Visual Entailment is examined in
Section~\ref{sec:experiments}.
 
\subsection{Vietnamese Multimodal Resources}
\label{subsec:vietnamese}
 
\subsubsection{Image Captioning}
 
Vietnamese multimodal research has its deepest roots in image captioning, and the available captioning datasets provide both the image sources and annotation precedents
upon which OpenViVE builds. UIT-ViIC~\cite{lam2020uitviic} introduced the first Vietnamese image captioning dataset with controlled scenes, establishing annotation guidelines for Vietnamese visual description that balance linguistic
naturalness with visual accuracy. UIT-OpenViIC~\cite{bui2023uitopenviic} the image source for OpenViVE extends this to open-domain scenes capturing daily
life, cultural activities, street scenes, and local landscapes in Vietnam, with annotations collected under strict inter-annotator agreement requirements. The
open-domain nature of UIT-OpenViIC is particularly important for OpenViVE: it ensures that premise images reflect authentic Vietnamese visual contexts, including culturally specific objects (\textit{phở}, \textit{bánh mì},
\textit{quẩy}, and \textit{áo dài}), traditional architectural styles, and social scenarios that are
absent from English-centric image collections such as Flickr30k or MS-COCO. KTVIC~\cite{pham2024ktvic} contributes a life-domain Vietnamese image captioning
dataset in which each image is paired with multiple captions, enabling evaluation of
caption diversity and reference-based metrics. ViTextCaps~\cite{nguyen2026linguistically}
addresses the more technically demanding task of text-rich image captioning in
Vietnamese, requiring models to integrate scene-text recognition, OCR processing, and
visual captioning into a unified generation pipeline.
 
These captioning datasets collectively provide a strong foundation for Vietnamese
image--text understanding, but they are designed exclusively to evaluate the quality
of \emph{generated} descriptions. They provide no negative (contradiction) or
uncertain (neutral) hypotheses, and they do not require models to classify a logical
semantic relation. Captioning performance, therefore, cannot serve as a proxy for
VE ability.
 
\subsubsection{Visual Question Answering}
 
Vietnamese VQA resources have diversified considerably, spanning open ended answer generation, scene text comprehension, and multi task reasoning. OpenViVQA~\cite{nguyen2023openvivqa} introduces the first large scale open ended Vietnamese VQA dataset, where models generate natural language answers grounded in everyday Vietnamese visual scenes rather than selecting from a fixed answer set. ViTextVQA~\cite{nguyen2024vitextvqa} targets text based VQA in Vietnamese, requiring models to read and interpret Vietnamese text embedded in images, such as street signs, product labels, and menus, while integrating scene text recognition with visual grounding. ViOCRVQA~\cite{pham2024viocrvqa} further focuses on OCR centric VQA, emphasising the role of character level text recognition in multimodal question answering for Vietnamese. VMMU~\cite{dang2025vmmu} provides a comprehensive evaluation of vision language models on Vietnamese multimodal reasoning across scene understanding, cultural knowledge, text recognition, and spatial reasoning. This evaluation reveals that current VLMs face substantial difficulties when Vietnamese language understanding and visual reasoning are jointly required. ViMACSA~\cite{nguyen2025new} extends the scope of Vietnamese multimodal research into fine grained opinion mining, introducing a benchmark for Multimodal Aspect Category Sentiment Analysis.
 
Despite this diversity, existing Vietnamese VQA benchmarks primarily frame multimodal inference as answer retrieval or generation rather than propositional relation verification. While VQA focuses on extracting or predicting a discrete answer conditioned on a targeted question, Visual Entailment requires evaluating the comprehensive truth value of an entire declarative hypothesis against visual evidence. Crucially, VE demands joint verification over the full compositional scope of a statement—including quantifiers, attribute modifiers, negation cues, and spatial relations. Consequently, high performance on entity- or text-centric VQA benchmarks such as ViTextVQA~\cite{nguyen2024vitextvqa} does not inherently translate to robust visual entailment reasoning, as the two tasks test fundamentally distinct dimensions of multimodal comprehension.
 
\subsubsection{Textual Natural Language Inference}
 
Textual NLI, the unimodal precursor to VE, has developed substantially in Vietnamese and provides important baselines and lessons for multimodal extension. ViNLI~\cite{huynh-etal-2022-vinli} introduced the first large scale Vietnamese NLI corpus with open domain premise and hypothesis pairs, establishing that Vietnamese textual inference is a non trivial task even without the visual modality. ViANLI~\cite{van2026new} extends Vietnamese NLI to the adversarial setting, applying a human and model in the loop (HAMIL) construction procedure to generate hypotheses that fool current models while remaining valid for human judges. This adversarial methodology is directly relevant to OpenViVE: it demonstrates that Vietnamese NLI models are vulnerable to hypothesis only exploitation, precisely the shortcut learning risk that motivates our PMI based artifact mitigation. ViLegalNLI~\cite{duong2026vilegalnli} addresses domain specific NLI in the Vietnamese legal domain, demonstrating that inference in specialised registers requires domain adapted representations.
 
The existence of mature Vietnamese textual NLI resources strengthens the case for OpenViVE in two ways. First, it confirms that Vietnamese hypothesis classification is a well defined problem at the text only level. Consequently, the challenge of OpenViVE is specifically the addition of visual grounding, meaning that models cannot fall back on language only inference strategies. Second, the ViANLI finding that Vietnamese NLI models are vulnerable to hypothesis only shortcuts directly motivates the artifact mitigation procedures in our construction pipeline.
 
Table~\ref{tab:viet_comparison} provides a structured comparison of Vietnamese multimodal resources along dimensions relevant to visual entailment evaluation. The defining intersection, which requires both three way entailment labels \emph{and} visual grounding, is occupied exclusively by OpenViVE. This confirms that no existing resource can substitute for it.
 
 
\begin{table*}[t]
  \centering
  \caption{Comparison of Vietnamese language and multimodal resources relevant
  to Visual Entailment. ``3-way'' indicates the use of
  Entailment/Contradiction/Neutral labels. ``Image'' indicates whether visual
  input is included in the task. ``Cross-modal reasoning'' indicates whether
  the task requires joint reasoning over visual and textual information.
  \cmark~= yes; \xmark~= no; Partial~= partial.}
  \label{tab:viet_comparison}
  \renewcommand{\arraystretch}{1.25}
  \setlength{\tabcolsep}{4.5pt}
  \resizebox{\linewidth}{!}{%
  \begin{tabular}{llp{1.5cm}ccccp{2.0cm}}
    \toprule
    \textbf{Dataset} & \textbf{Task} & \textbf{Scale (samples)}
      & \textbf{3-way} & \textbf{Image}
      & \textbf{Cross-modal Reasoning} & \textbf{Annotation Method} \\
    \midrule
    UIT-ViIC~\cite{lam2020uitviic}
      & Captioning & 19,250
      & \xmark & \cmark & \xmark & Handicraft \\
    UIT-OpenViIC~\cite{bui2023uitopenviic}
      & Captioning & 61,241
      & \xmark & \cmark & \xmark & Handicraft \\
    KTVIC~\cite{pham2024ktvic}
      & Captioning & 21,635
      & \xmark & \cmark & \xmark & Handicraft \\
    ViTextCaps~\cite{nguyen2026linguistically}
      & Text-rich captioning & 74,970
      & \xmark & \cmark & \xmark & Handicraft \\
    OpenViVQA~\cite{nguyen2023openvivqa}
      & Open-ended VQA & 37,000
      & \xmark & \cmark & Partial & Handicraft \\
    ViTextVQA~\cite{nguyen2024vitextvqa}
      & Text-based VQA & 50,000
      & \xmark & \cmark & Partial & Handicraft \\
    ViOCRVQA~\cite{pham2024viocrvqa}
      & OCR-VQA & 120,000+
      & \xmark & \cmark & \xmark & Semi-Automatic \\
    VMMU~\cite{dang2025vmmu}
      & Multi-task VL & 2,548
      & \xmark & \cmark & Partial & Handicraft \\
    ViMACSA~\cite{nguyen2025new}
      & Multimodal sentiment analysis & 10,000
      & \xmark & \cmark & Partial & Handicraft \\
    \midrule
    ViNLI~\cite{huynh-etal-2022-vinli}
      & Textual NLI & 30,000
      & \cmark & \xmark & \xmark & Handicraft \\
    ViANLI~\cite{van2026new}
      & Adversarial NLI & 10,000
      & \cmark & \xmark & \xmark & Handicraft \\
    ViLegalNLI~\cite{duong2026vilegalnli}
      & Legal NLI & 42,012
      & \cmark & \xmark & \xmark & Handicraft \\
    \midrule
    \textbf{OpenViVE (Ours)}
      & \textbf{Visual Entailment} & \textbf{62,268}
      & \cmark & \cmark & \cmark & Semi-Automatic \\
    \bottomrule
  \end{tabular}
  }
\end{table*}
 
\subsection{LLM-Assisted Data Construction for Multimodal Learning}
\label{subsec:llm_construction}
 
\subsubsection{Generative Approaches to Vision--Language Data}
 
The high annotation cost of multimodal datasets has motivated a growing body of work
on generative and semi-automatic data construction. In the image generation direction,
Reijtenbach~et~al.~\cite{reijtenbach2025dataset} construct synthetic VE training
data by using Stable Diffusion to generate premise images from SNLI textual premises.
While this approach reduces annotation burden substantially, the resulting data quality
is limited by the faithfulness of image generation: the model frequently produces
compositional errors, unrealistic object arrangements, and visual artifacts that would
not appear in authentic photographs, and it systematically fails to generate culturally
specific visual content outside its training distribution. For Vietnamese cultural
content restaurants, street markets, Vietnamese
architectural styles this limitation is especially severe.
 
In the text generation direction, the use of large language models to produce
multimodal supervision signals has become increasingly common.
LLaVA~\cite{liu2023llava} demonstrates that GPT-4-generated multimodal
instruction-following data enables effective visual instruction tuning, showing that
the quality and diversity of generated instructions matter as much as their quantity.
ShareGPT4V~\cite{chen2023sharegpt4v} extends this finding to image captioning,
demonstrating that GPT-4V-generated captions improve image-text alignment and
downstream multimodal performance across multiple benchmarks, emphasising that data
quality is at least as important as data scale. These results motivate the use of
capable LLMs as hypothesis generators in OpenViVE: if a sufficiently capable
model can generate diverse, semantically accurate, and linguistically natural hypotheses
given label-specific constraints, the resulting data can approach expert human
annotation quality at a fraction of the cost. However, LLM-generated text may contain
hallucinated details, repetitive stylistic patterns, or label-correlated lexical cues
that allow hypothesis-only classifiers to exploit the benchmark without attending to
the image.
 
\subsubsection{Annotation Artifacts in NLI and VE Data}
 
A persistent risk in large-scale NLI and VE data construction is the introduction of
\emph{annotation artifacts}-systematic statistical associations between surface
features of the hypothesis text and the assigned label that allow models to predict
the correct label without attending to the visual premise.
Gururangan~et~al.~\cite{gururangan2018annotation} formally characterise this problem
for textual NLI: crowdworkers tasked with generating hypotheses for specific labels
develop systematic lexical patterns that serve as label cues, including the use of
negation words for Contradiction labels, hedge phrases (\textit{might}, \textit{perhaps})
for Neutral labels, and concrete action verbs for Entailment labels. Models trained on
artifact-contaminated datasets achieve high test accuracy through hypothesis-only
inference but fail to generalise to artifact-controlled evaluation sets.
 
Subsequent work has proposed several strategies for reducing annotation artifacts. Adversarial NLI ~\cite{van2026new} applies a human-and-model-in-the-loop (HAMIL) construction procedure in which annotators must write hypotheses that fool current models while remaining valid for human judges, producing a dataset that is systematically more difficult and less artifact-prone than standard NLI benchmarks.
Dataset cartography ~\cite{swayamdipta2020dataset} provides a diagnostic framework
for mapping dataset examples by their training dynamics whether examples are
consistently correct, ambiguous, or consistently incorrect for a trained model enabling
targeted curation of the most informative and least artifact-prone training examples.
These methods demonstrate that artifact mitigation requires active intervention in the
data construction process, not merely post-hoc filtering.

In the VE setting, the artifact risk is compounded. A VE dataset constructed with LLM
assistance must control not only for lexical artifacts in the hypothesis text but also
for semantic misalignments between the hypothesis and the image, hallucinated details
generated without visual grounding, and label-specific generation styles that differ
systematically across the three classes. Failing to address any of these risks produces
a benchmark in which models can achieve high accuracy through a combination of
language-only shortcuts and image--text mismatches, without performing genuine
visual-textual reasoning.
 
\subsubsection{The OpenViVE Construction Strategy}
\label{subsubsec:openvive_strategy}
 
OpenViVE addresses the risks identified above through a three-layer mitigation
strategy that advances beyond the approaches used in prior VE dataset construction.
 
The first layer is \textbf{rule-constrained LLM generation}. Rather than prompting
Gemini 2.5 Flash to generate hypotheses freely, we associate each label with a distinct
set of generation rules (see Table~\ref{tab:generation_rules} in Section~\ref{subsec:prompt_design}) that
constrain the semantic type of modification permitted for each label. Entailment
hypotheses must describe information directly supported by observable visual content.
Contradiction hypotheses must introduce controlled semantic conflicts changes in object
identity, quantity, attribute, action, or event rather than arbitrary negations.
Neutral hypotheses must introduce plausible but visually unverifiable
information intentions, temporal contexts, emotional states, background
events rather than implausible fabrications. This rule structure prevents the LLM
from developing free-form label-specific generation styles and reduces systematic
lexical differences across label classes.
 
The second layer is \textbf{iterative human-in-the-loop prompt validation}. Generated
hypotheses are reviewed by trained native Vietnamese annotators before large-scale
generation proceeds. Prompt versions that produce frequent hallucinations, unnatural
Vietnamese expressions, or label confusion are revised and re-evaluated. Only prompts
achieving inter-annotator agreement above Fleiss' $\kappa = 0.85$ across 100 sampled
hypotheses are used for large-scale generation, ensuring that generation quality is
validated before it is scaled up.
 
The third layer is \textbf{PMI-based lexical artifact analysis and targeted rewriting}.
After large-scale generation, we compute the Pointwise Mutual Information (PMI) between
each hypothesis token and each label, identifying tokens that are overly predictive of
specific classes. Hypotheses containing high-PMI tokens are then manually revised to
reduce their hypothesis-only predictability while preserving semantic accuracy. The
effectiveness of this mitigation is verified empirically: before artifact reduction, a
hypothesis-only classifier achieves 0.96 accuracy on Entailment, 0.96 on Contradiction,
and 0.98 on Neutral, confirming the presence of strong label-predictive cues. After
artifact reduction, these figures drop to 0.80, 0.66, and 0.92 respectively,
demonstrating that the mitigation procedure substantially reduces shortcut-learning
opportunities while preserving the semantic structure of the dataset.

Table~\ref{tab:construction_comparison} compares the data construction approaches used
in OpenViVE and related multimodal datasets, highlighting the dimensions along
which OpenViVE's hybrid pipeline advances beyond existing practices.
OpenViVE is the only work in this comparison to employ both explicit artifact
\emph{analysis} and explicit artifact \emph{mitigation} while targeting a low-resource
language with real-world culturally grounded images.
 
 
\begin{table*}[t]
  \centering
  \caption{Comparison of LLM-assisted and semi-automatic multimodal data construction
           approaches.
           ``LLM hyp.'' = LLM used to generate hypotheses or instructions.
           ``Real imgs.'' = real-world photographs used as premises (vs.\ generated).
           ``Human val.'' = human validation of generated content.
           ``Art.\ anal.'' = explicit quantitative analysis of lexical artifacts.
           ``Art.\ mitig.'' = active steps taken to reduce label-predictive cues.
           ``Low-res.'' = dataset targets a low-resource language.
           \cmark~= yes; \xmark~= no; Partial~= partial.}
  \label{tab:construction_comparison}
  \renewcommand{\arraystretch}{1.25}
  \setlength{\tabcolsep}{4pt}
  \resizebox{\linewidth}{!}{%
  \begin{tabular}{lcccccc}
    \toprule
    \textbf{Work}
      & \textbf{LLM hyp.}
      & \textbf{Real imgs.}
      & \textbf{Human val.}
      & \textbf{Art.\ anal.}
      & \textbf{Art.\ mitig.}
      & \textbf{Low-res.} \\
    \midrule
    SNLI-VE~\cite{xie2019visual}
      & \xmark & \cmark & Partial & \xmark & \xmark & \xmark \\
    e-SNLI-VE~\cite{do2020snli}
      & \xmark & \cmark & \cmark & Partial & Manual corr. & \xmark \\
    LLaVA~\cite{liu2023llava}
      & \cmark~(GPT-4) & \cmark & \xmark & \xmark & \xmark & \xmark \\
    ShareGPT4V~\cite{chen2023sharegpt4v}
      & \cmark~(GPT-4V) & \cmark & \xmark & \xmark & \xmark & \xmark \\
    Reijtenbach~et~al.~\cite{reijtenbach2025dataset}
      & \xmark & \xmark & \xmark & \xmark & \xmark & \xmark \\
    Defeasible VE~\cite{zhang2025defeasible}
      & Partial & \cmark & Partial & \xmark & \xmark & \xmark \\
    ViANLI~\cite{van2026new}
      & \xmark & --- & \cmark & Partial & HAMIL & \cmark \\
    \midrule
    \textbf{OpenViVE (ours)}
      & \cmark~(Gemini 2.5) & \cmark & \cmark
      & \cmark~(PMI) & \cmark~(PMI\,+\,rewrite) & \cmark \\
    \bottomrule
  \end{tabular}
  }
\end{table*}

\section{Benchmark Dataset Creation}
\label{sec:dataset_creation}

\subsection{Task Definition}
\label{subsec:task_definition}
Visual Entailment (VE)~\cite{xie2019visual} is formulated as a multimodal
inference task that determines the semantic relation between an image and a
textual hypothesis. Given an image $I$ as the visual premise and a Vietnamese
sentence $H$ as the hypothesis, the task is to predict one of three labels:
\textit{Entailment}, \textit{Contradiction}, or \textit{Neutral}. Formally,
given an image--hypothesis pair $(I,H)$, the target label
$y \in \{E,C,N\}$ indicates whether the hypothesis is supported by the
visible evidence in the image, contradicted by that evidence, or cannot be
verified or refuted from the image alone.

In OpenViVE, premise images are drawn from UIT-OpenViIC~
\cite{bui2023uitopenviic}, an open-domain Vietnamese image-captioning
resource containing real-world scenes related to daily life and culture in
Vietnam. Vietnamese hypotheses are constructed under label-specific
constraints described in Section~\ref{subsec:dataset_creation}. OpenViVE is
therefore designed to evaluate grounded Vietnamese image--text inference,
while the extent to which textual shortcuts remain exploitable is examined
separately through lexical-artifact and hypothesis-only analyses.

\subsection{Dataset Creation}
\label{subsec:dataset_creation}

\begin{figure}[!htbp]
\centering
\includegraphics[width=1\linewidth]{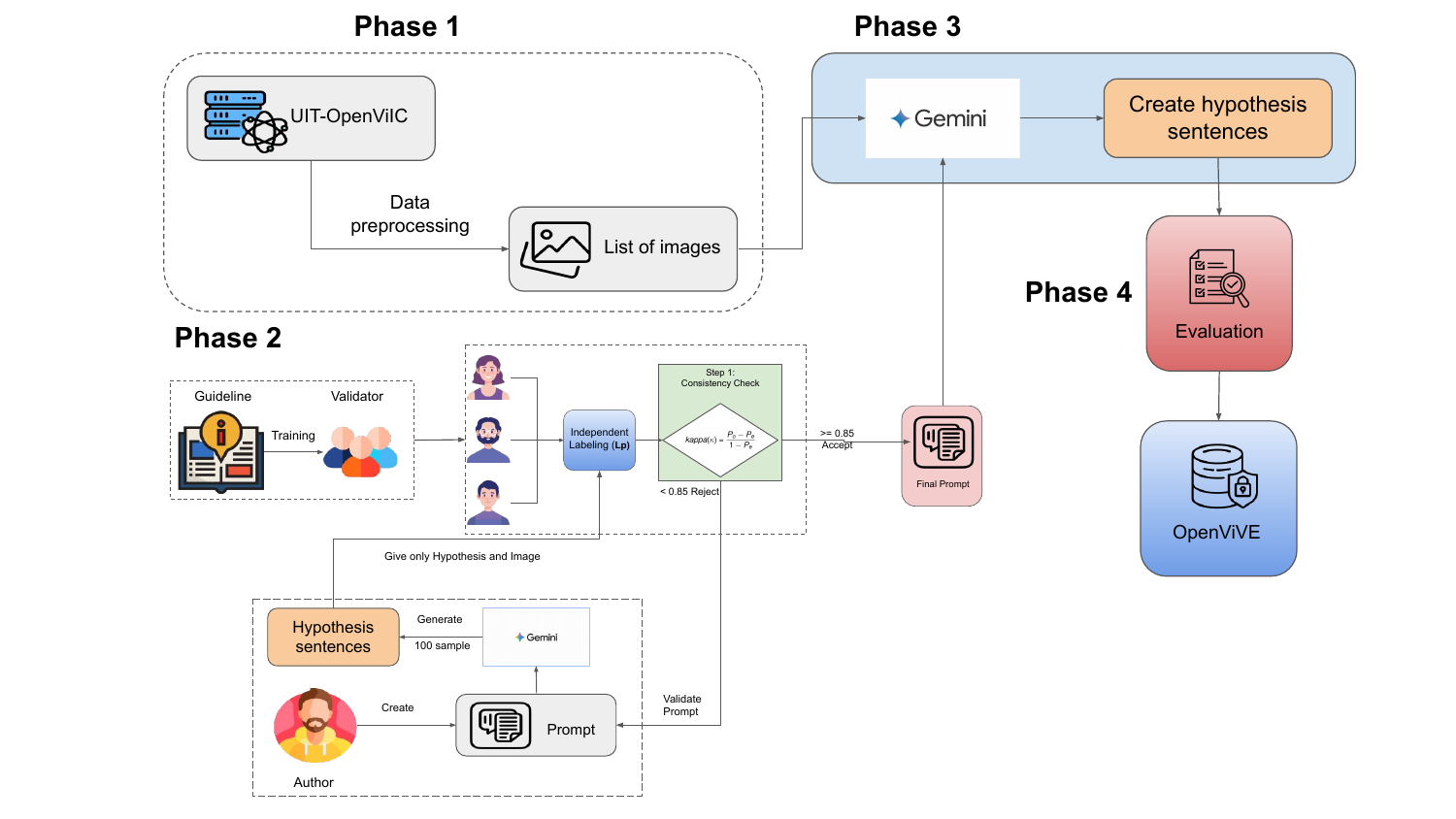}
\caption{Overview of the OpenViVE dataset construction pipeline.}
\label{fig:data_pipeline}
\end{figure}

OpenViVE is constructed through a four-stage semi-automatic pipeline, as
illustrated in Figure~\ref{fig:data_pipeline}. The pipeline consists of:
(1) selecting visually suitable premise images from UIT-OpenViIC;
(2) developing label-specific generation prompts through validator-guided
iteration; (3) generating Vietnamese hypotheses at scale; and
(4) validating the generated image--hypothesis pairs and refining the corpus
before final release. The overall design combines model-assisted generation
for scalability with human validation and post-generation diagnostics for
semantic and corpus-level quality control.

\subsubsection{Phase 1: Data Source Selection}
\label{subsec:data_source}

OpenViVE uses UIT-OpenViIC~\cite{bui2023uitopenviic} as the source of its
visual premises. The corpus contains 13,100 open-domain images collected
primarily using Vietnamese search queries and paired with manually written
Vietnamese captions, covering a broad range of everyday and
Vietnamese-context scenes. From this collection, we further screen out
images that are unsuitable for reliable hypothesis construction and
validation, including cases that are severely blurred, heavily occluded, or
contain insufficient visible information to support meaningful
image--hypothesis relations. After this screening step, 10,378 images are
retained. The resulting subset preserves the broad open-domain and
Vietnamese-context coverage of the source corpus while removing visually
uninformative cases.

\subsubsection{Phase 2: Validator Training and Prompt Design}
\label{subsec:prompt_design}

A pool of student validators was recruited from leading universities in Vietnam. All validators were native Vietnamese speakers with strong language proficiency, sufficient background knowledge, and the ability to jointly interpret visual and textual information. Since OpenViVE is designed for multimodal reasoning, validators served as an important quality-control component during prompt development by assessing whether the generated hypotheses were semantically aligned with the corresponding images and intended labels.

To establish a consistent generation and validation standard, we prepared
guidelines covering the task definition, label semantics, and validation
criteria. Following the three-way Visual Entailment formulation of
SNLI-VE~\cite{xie2019visual}, an \textit{Entailment} hypothesis must be
directly supported by visible evidence in the image, whereas a
\textit{Contradiction} hypothesis must conflict with observable visual
content, typically through a change in object identity, quantity, attribute,
action, or event. A \textit{Neutral} hypothesis must remain plausible in the
image context while containing information that cannot be verified or
refuted from the image alone.

The generation prompts were written in Vietnamese and organized according
to the three Visual Entailment labels. Each label was associated with a
distinct set of generation rules, as summarized in
Table~\ref{tab:generation_rules}. For \textit{Entailment}, the prompt guided
the language model to generate hypotheses grounded in observable visual
evidence. For \textit{Contradiction}, visually supported statements were
used as the basis for constructing minimally modified hypotheses that
introduced controlled semantic conflicts. For \textit{Neutral}, the prompt
encouraged plausible statements containing contextual information that
could not be confirmed or refuted from the image alone. The complete
label-specific prompt templates used for hypothesis generation are provided
in Appendix~\ref{app:generation_prompts}.
\begin{table}[t]
\centering
\small
\setlength{\tabcolsep}{5pt}
\renewcommand{\arraystretch}{1.15}
\caption{Label-specific rules used for hypothesis generation in OpenViVE.}
\label{tab:generation_rules}
\begin{tabular}{p{0.18\columnwidth} p{0.75\columnwidth}}
\toprule
\textbf{Label} & \textbf{Generation rules} \\
\midrule

Entailment &
\begin{minipage}[t]{\linewidth}
\begin{enumerate}[leftmargin=*, nosep]
    \item Infer from visible actions.
    \item Use synonymous expressions.
    \item Identify visible entities, locations, or brands.
    \item Use natural Vietnamese phrasing.
\end{enumerate}
\end{minipage}
\\

\midrule

Contradiction &
\begin{minipage}[t]{\linewidth}
\begin{enumerate}[leftmargin=*, nosep]
    \item Replace with an opposite or incompatible expression.
    \item Change the quantity.
    \item Change a supported time reference.
    \item Make an incorrect claim about an object.
    \item Make an incorrect claim about an event.
\end{enumerate}
\end{minipage}
\\

\midrule

Neutral &
\begin{minipage}[t]{\linewidth}
\begin{enumerate}[leftmargin=*, nosep]
    \item Add an unverifiable time detail.
    \item Add an unverifiable gender detail.
    \item Add an unverifiable emotion.
    \item Add an unverifiable size detail.
    \item Add a related past or future event.
    \item Add a plausible purpose or intention.
\end{enumerate}
\end{minipage}
\\

\bottomrule
\end{tabular}
\end{table}

Prompt development was conducted iteratively through pilot validation. For
each prompt version, 100 generated image--hypothesis pairs were sampled and
independently reviewed by validators. Each sample was evaluated with respect
to the intended VE label, linguistic naturalness in Vietnamese, and semantic
clarity relative to the corresponding image. Validators also recorded error
feedback for problematic cases, such as hallucinated information, ambiguous
image--text relations, unnatural expressions, or label inconsistencies.

Inter-annotator agreement on the label evaluation was measured using
Fleiss' $\kappa$~\cite{fleiss1971measuring}. Given the mean observed
agreement $\bar{P}$ and the agreement expected by chance $\bar{P}_e$,
Fleiss' $\kappa$ is defined as:
\begin{equation}
\kappa = \frac{\bar{P} - \bar{P}_e}{1 - \bar{P}_e}.
\end{equation}
A prompt version was accepted only when the resulting agreement exceeded
the predefined threshold of $\kappa > 0.85$.

Prompt versions associated with recurring validation errors were revised
based on the collected feedback and evaluated again using the same procedure.
This iterative refinement was conducted over five rounds. The final prompt
set was retained after satisfying the predefined agreement criterion and was
subsequently used for large-scale hypothesis generation.

\subsubsection{Phase 3: Large-scale Hypothesis Generation}
\label{subsec:hypothesis_generation}

After the prompt set was finalized, large-scale hypothesis generation was
performed using Gemini 2.5 Flash. The decoding parameters were set to a
temperature of 0.4, top-$p$ of 0.8, and top-$k$ of 20, and the same
generation configuration was used throughout the construction process.

For each retained image, hypotheses were generated separately using the
label-specific prompt templates developed in Phase 2. Two hypotheses were
generated for each of the three VE labels, yielding two Entailment, two
Contradiction, and two Neutral hypotheses per image. Each visual premise
therefore corresponds to six image--hypothesis pairs.

With 10,378 retained images, the generation process produced 62,268
image--hypothesis pairs in total, including 20,756 pairs for each label.
The resulting corpus is therefore balanced across Entailment, Contradiction,
and Neutral. The dataset was subsequently partitioned into training,
development, and test sets at the image level so that all hypotheses
associated with the same image remain within a single split, preventing
visual overlap across partitions. Detailed split statistics are reported in
Table~\ref{tab:dataset_statistics}.

\subsubsection{Phase 4: Dataset Validation}
\label{subsec:dataset_validation}

To ensure the reliability of OpenViVE, a final validation stage was conducted
on the development and test splits. Three trained validators independently
reviewed each image--hypothesis pair without access to one another's
decisions. Each sample was evaluated according to three criteria:
(1) consistency between the hypothesis, the image, and the assigned VE label;
(2) grammatical correctness and lexical appropriateness in Vietnamese; and
(3) semantic clarity with respect to the image.

For each criterion, validators provided a binary judgment, where
\textit{Yes} indicates that the sample satisfies the criterion and
\textit{No} indicates that it does not. A validator accepted a sample only
when all three criteria were satisfied. The final decision was then determined
by majority voting across the three validators. A sample was considered valid
when at least two of the three validators accepted it. Samples that did not
satisfy this majority criterion were rewritten and subsequently re-evaluated
before being included in the final dataset.

As shown in Table~\ref{tab:openvive_validation}, the initial validation pass
accepted 5,756 of 6,228 development samples and 5,787 of 6,228 test samples,
corresponding to acceptance rates of 0.92 and 0.93, respectively.
Inter-annotator agreement reached Fleiss' $\kappa$ values of 0.91 on the
development split and 0.92 on the test split. Samples rejected during the
initial validation pass were rewritten and re-evaluated so that the final
development and test sets each retained 6,228 validated image--hypothesis
pairs.

\begin{table}[t]
\centering
\small
\setlength{\tabcolsep}{5pt}
\renewcommand{\arraystretch}{1.15}
\caption{Initial validation outcomes on the OpenViVE development and test splits.}
\label{tab:openvive_validation}
\begin{tabular*}{\columnwidth}{@{\extracolsep{\fill}} l c c c c}
\toprule
\textbf{Split}
& \textbf{Total Samples}
& \textbf{Accepted Samples}
& \textbf{Acceptance Rate}
& \textbf{Fleiss' $\kappa$} \\
\midrule
Dev  & 6,228 & 5,756 & 0.92 & 0.91 \\
Test & 6,228 & 5,787 & 0.93 & 0.92 \\
\bottomrule
\end{tabular*}
\end{table}

In parallel with human validation, PMI-based analysis was used to identify
label-specific lexical bias and overly predictive terms. When such artifacts
were detected, the corresponding hypotheses were revised to improve lexical
diversity and reduce annotation bias. Together, human validation and
PMI-based refinement provide complementary quality-control mechanisms for
improving semantic consistency and reducing exploitable textual artifacts in
the final corpus.

\subsection{Dataset Analysis}
\label{subsec:dataset_analysis}

In this subsection, we analyze OpenViVE from multiple perspectives to better understand its scale, balance, and linguistic characteristics. We first present the overall dataset statistics, including the number of images, image--hypothesis pairs, and label distribution across the training, development, and test splits. We then examine the length distribution of Vietnamese hypotheses to assess whether the generated sentences remain consistent across splits. Finally, we analyze lexical artifacts to identify potential label-specific cues that may lead models to rely on hypothesis-only shortcuts instead of grounded visual-textual reasoning.

\subsubsection{Initial Statistics}
\label{subsubsec:initial_statistics}

Table~\ref{tab:dataset_statistics} presents the overall statistics of OpenViVE. The dataset contains 10,378 images and 62,268 image--hypothesis pairs. Each image is paired with six Vietnamese hypotheses, including two hypotheses for each of the three labels: \textit{Entailment}, \textit{Contradiction}, and \textit{Neutral}. The data are split into training, development, and test sets following an 80/10/10 ratio.

The dataset is split at the image level, meaning that all hypotheses associated with the same image are assigned to the same subset. This strategy prevents the same visual premise from appearing in multiple splits and reduces the risk of visual leakage during evaluation. As shown in Table~\ref{tab:dataset_statistics}, the label distribution is fully balanced in all splits. Each label accounts for one third of the dataset, with 20,756 samples for \textit{Entailment}, 20,756 samples for \textit{Contradiction}, and 20,756 samples for \textit{Neutral}. This balanced design helps reduce label-frequency bias and makes OpenViVE suitable for evaluating visual-textual reasoning rather than prior label distribution.

\begin{table}[!t]
\centering
\small
\renewcommand{\arraystretch}{1.15}
\caption{Statistics of OpenViVE across data splits.}
\label{tab:dataset_statistics}
\begin{tabular*}{\columnwidth}{@{\extracolsep{\fill}}lrrrrr@{}}
\toprule
\textbf{Split} & \textbf{Images} & \textbf{Entailment} & \textbf{Contradiction} & \textbf{Neutral} & \textbf{Total} \\
\midrule
Train & 8,302  & 16,604 & 16,604 & 16,604 & 49,812 \\
Dev   & 1,038  & 2,076  & 2,076  & 2,076  & 6,228  \\
Test  & 1,038  & 2,076  & 2,076  & 2,076  & 6,228  \\
\midrule
All   & 10,378 & 20,756 & 20,756 & 20,756 & 62,268 \\
\bottomrule
\end{tabular*}
\end{table}

\subsubsection{Length Distribution}
\label{subsubsec:length_distribution}

\begin{figure}[!htbp]
\centering
\includegraphics[width=\linewidth]{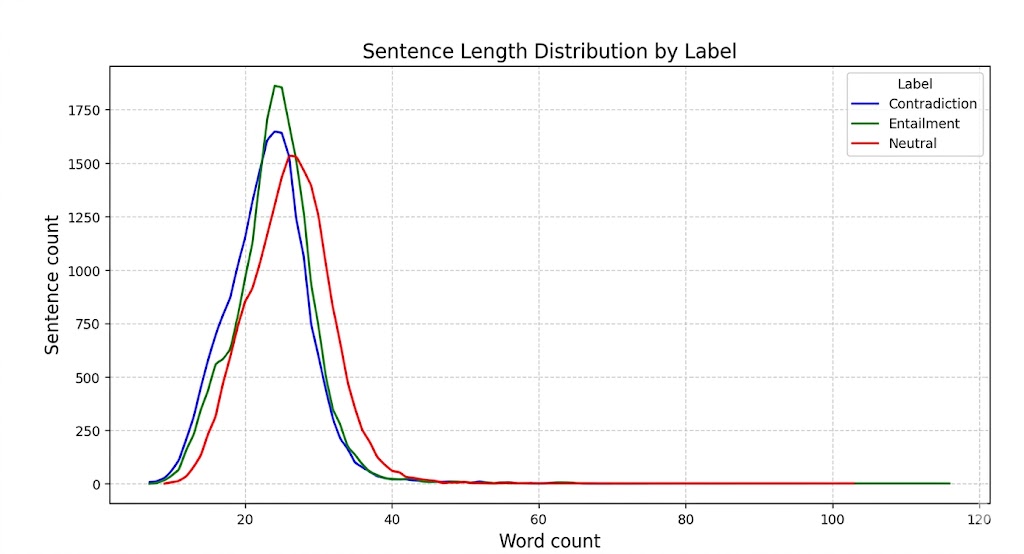}
\caption{Hypothesis-length distribution of Vietnamese hypotheses by label
in OpenViVE.}
\label{fig:length_distribution}
\end{figure}

We further analyze the length distribution of Vietnamese hypotheses in
OpenViVE. Hypothesis length is an important surface-level characteristic,
as substantial differences across labels or data splits may introduce
unintended distribution shifts or provide shallow textual cues. 
Table~\ref{tab:length_statistics} reports the mean, median, mode, maximum
length, and vocabulary size across the training, development, and test
splits. Figure~\ref{fig:length_distribution} further illustrates the
hypothesis-length distributions across the three VE labels.

\begin{table}[!htbp]
\centering
\small
\renewcommand{\arraystretch}{1.15}
\caption{Hypothesis length and vocabulary statistics across data splits.}
\label{tab:length_statistics}
\begin{tabular*}{\columnwidth}{@{\extracolsep{\fill}}lrrrrr@{}}
\toprule
\textbf{Split}
& \textbf{Mean}
& \textbf{Median}
& \textbf{Mode}
& \textbf{Max}
& \textbf{Vocab. size} \\
\midrule
Train & 23.6 & 23.0 & 24.0 & 94.0  & 6,544 \\
Dev   & 23.1 & 23.0 & 22.0 & 80.0  & 3,279 \\
Test  & 23.2 & 23.0 & 23.0 & 114.0 & 3,343 \\
\bottomrule
\end{tabular*}
\end{table}

As shown in Table~\ref{tab:length_statistics}, the hypothesis-length
statistics are broadly consistent across the three data splits. The mean
length ranges from 23.1 to 23.6, while the median remains 23.0 across the
training, development, and test sets. The mode also varies only slightly,
from 22.0 to 24.0. These results indicate that the data partitioning does
not introduce substantial differences in hypothesis length across splits.

The training split contains the largest vocabulary because it accounts for
the majority of the dataset, while the development and test splits contain
smaller but comparable vocabulary sizes. This pattern is consistent with
their relative dataset sizes and suggests that the two evaluation splits
maintain similar lexical coverage.

As shown in Figure~\ref{fig:length_distribution}, the three labels exhibit
substantially overlapping length distributions, with most hypotheses
concentrated between approximately 20 and 30 words. Entailment and
Contradiction show particularly similar distributions, which is consistent
with the construction process in which contradiction hypotheses are often
formed by modifying visually grounded statements. Neutral hypotheses are
slightly shifted toward longer lengths, reflecting the addition of plausible
but visually unverifiable information such as temporal context, emotion,
intention, or background events.

Overall, the substantial overlap among the three distributions suggests that
hypothesis length does not provide a strong label-specific cue at the
distribution level. Together with the similar statistics across data splits,
these results indicate that OpenViVE maintains broadly consistent
surface-level length characteristics across its training and evaluation sets.
\subsubsection{Part-of-Speech Distribution}
\label{subsubsec:pos_distribution}

To better understand word usage patterns in OpenViVE, we analyze the part-of-speech (POS) distribution of Vietnamese hypotheses across the three labels. Since Vietnamese text requires word segmentation before downstream linguistic analysis, we first apply VnCoreNLP~\cite{vu2018vncorenlp} for Vietnamese word segmentation, and then use the Underthesea toolkit for POS tagging. For clarity, the original POS tags are grouped into six main categories: Nouns, Verbs, Adjectives, Prepositions, Adjuncts, and Others.

\begin{figure}[!htbp]
\centering
\includegraphics[width=\linewidth]{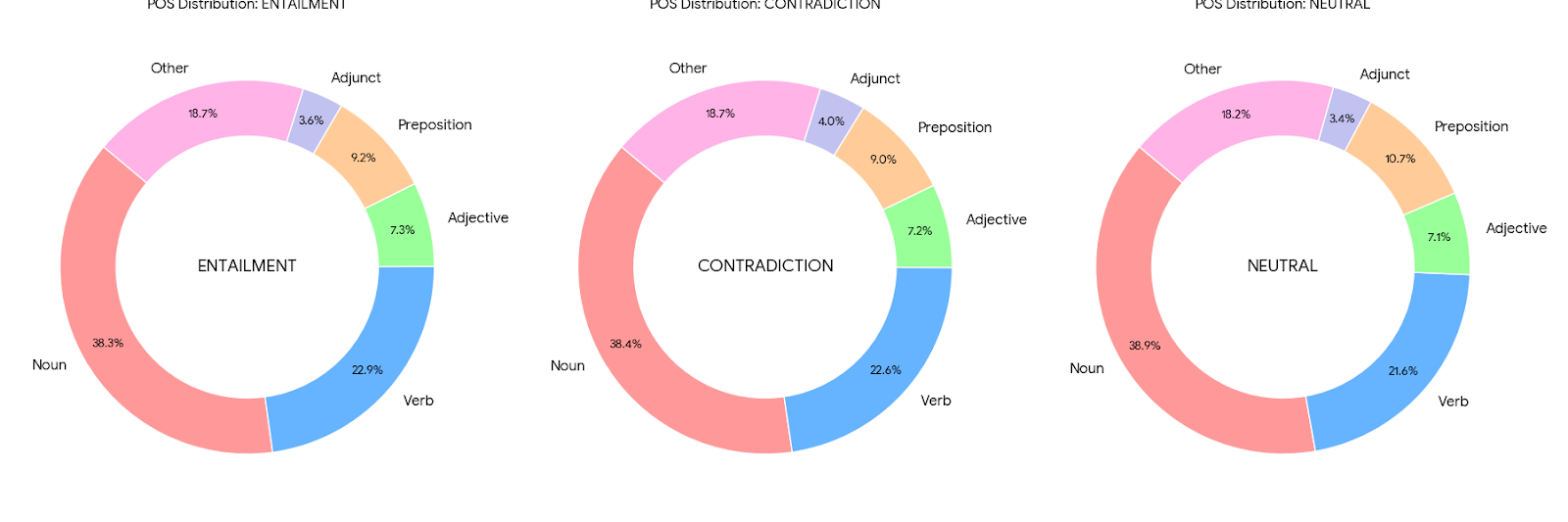}
\caption{Part-of-speech distribution of Vietnamese hypotheses across the three labels in OpenViVE.}
\label{fig:pos_distribution}
\end{figure}

As shown in Figure~\ref{fig:pos_distribution}, the POS distributions are highly consistent across \textit{Entailment}, \textit{Contradiction}, and \textit{Neutral}. Nouns and verbs account for the largest proportions in all three labels, reflecting the object- and action-oriented nature of visual entailment hypotheses. This pattern is reasonable because hypotheses in OpenViVE often describe visible entities, people, objects, actions, and events in the image. In contrast, adjectives and adjuncts appear less frequently, suggesting that the dataset does not rely heavily on descriptive modifiers or shallow stylistic patterns.

The similarity between \textit{Entailment} and \textit{Contradiction} is also expected, as contradiction hypotheses are often created by minimally modifying visually grounded statements while preserving a similar sentence structure. Neutral hypotheses follow a comparable POS distribution, although they may include slightly more contextual expressions related to time, purpose, or unverifiable background information. Overall, the POS analysis suggests that OpenViVE does not exhibit strong label-specific grammatical bias at the coarse POS level. This reduces the possibility that models can predict labels based only on superficial syntactic patterns, and further supports the suitability of OpenViVE for evaluating grounded Vietnamese visual-textual reasoning.

\subsubsection{Artifact Analysis}
\label{subsubsec:artifact_analysis}

To detect potential annotation artifacts, we compute Pointwise Mutual Information (PMI) between hypothesis tokens and labels. Before computing PMI, Vietnamese hypotheses are segmented into tokens. Given a token $w$ and a label $y$, PMI is defined as:
\begin{equation}
\operatorname{PMI}(w,y) = \log \frac{P(w,y)}{P(w)P(y)}.
\end{equation}

A higher PMI value indicates that a token or phrase is more strongly associated with a specific label and may therefore serve as a shallow lexical cue. In OpenViVE, PMI is used as a diagnostic tool to identify label-discriminative expressions and guide artifact refinement. Table~\ref{tab:pmi_artifacts} presents representative residual Vietnamese tokens and phrases after the refinement process.

\begin{table}[!htbp]
\centering
\small
\renewcommand{\arraystretch}{1.15}
\caption{Representative residual label-discriminative Vietnamese tokens and phrases based on PMI.}
\label{tab:pmi_artifacts}
\begin{tabular*}{0.85\columnwidth}{@{\extracolsep{\fill}}llcc@{}}
\toprule
\textbf{Token} & \textbf{English gloss} & \textbf{Label} & \textbf{PMI} \\
\midrule
Mặc        & Wearing          & E & 1.52 \\
Nhiều      & Many             & E & 1.49 \\
Hàng       & Row/line         & E & 1.16 \\
Chậm rãi   & Slowly           & E & 0.92 \\
Gần gũi    & Close/familiar   & E & 0.89 \\
\midrule
Lạnh lẽo   & Cold/desolate    & C & 1.58 \\
Đơn điệu   & Monotonous       & C & 1.57 \\
Vắng tanh  & Completely empty & C & 1.56 \\
Trống trơn & Bare/empty       & C & 1.55 \\
Khô cằn    & Arid/barren      & C & 1.54 \\
\midrule
Dự định    & Intention/plan   & N & 1.58 \\
Dự kiến    & Expected/planned & N & 1.58 \\
Ước chừng  & Approximately    & N & 1.58 \\
Ngày mai   & Tomorrow         & N & 1.58 \\
14:00           & 2:00 PM          & N & 1.58 \\
\bottomrule
\end{tabular*}
\end{table}
PMI results reveal that some label-specific lexical signals remain after refinement, particularly for expressions that are naturally tied to the semantic construction of each label. To reduce the effect of such shortcuts, high-PMI tokens and phrases were reviewed under controlled rewriting constraints, ensuring that superficial lexical cues were minimized while the original image--hypothesis relation was preserved. We do not treat PMI as evidence that OpenViVE can be solved from text alone; rather, it serves as a corpus-level diagnostic for identifying residual shortcut cues. The remaining effect of these cues is further evaluated through the hypothesis-only analysis in Section~\ref{subsubsec:hypothesis_only_artifact}.

\section{Proposed Framework}
\label{sec:framework}
\begin{figure}[!htbp]
\centering
\includegraphics[width=\linewidth]{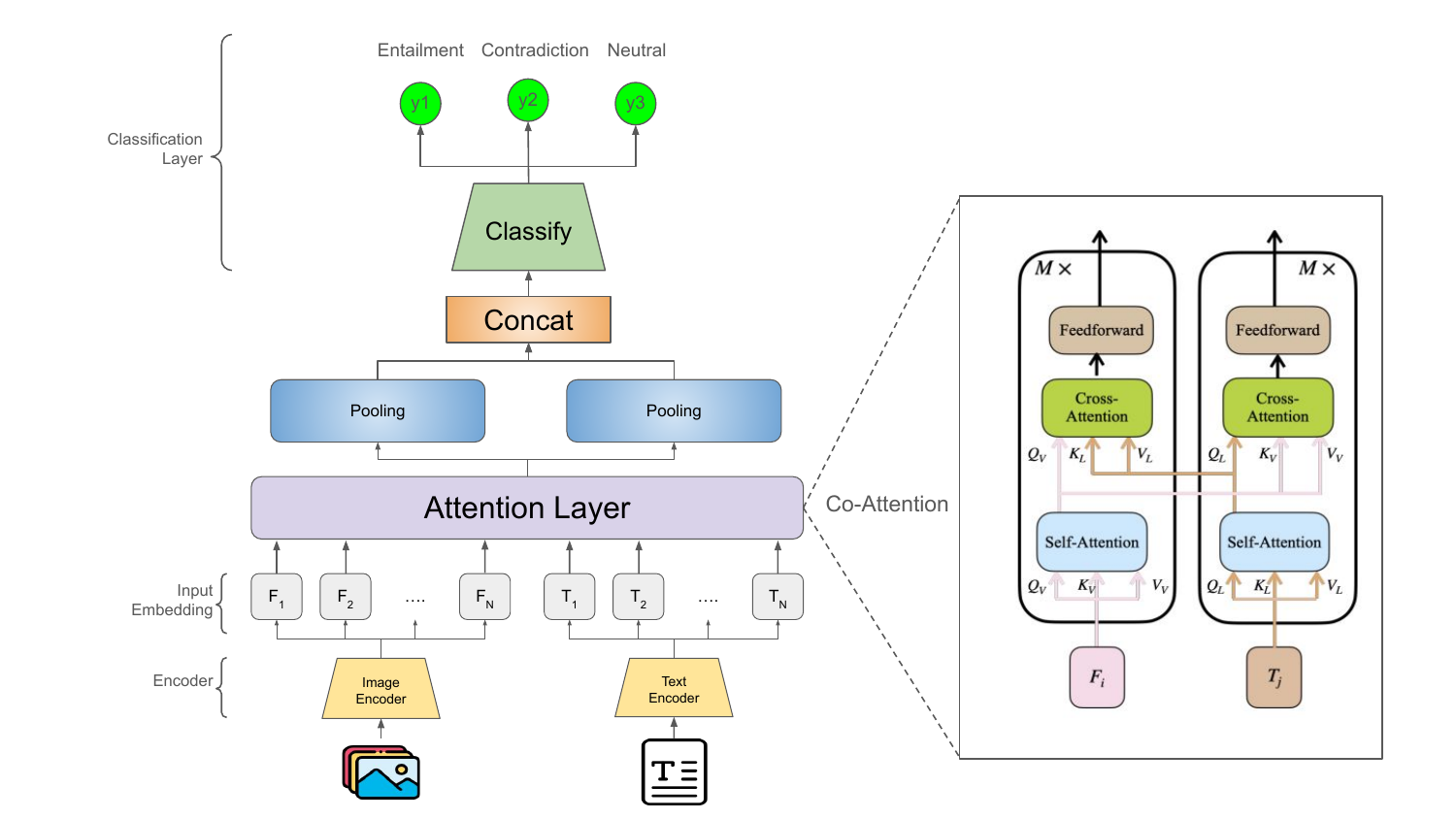}
\caption{Overview of the PhoBERT--ConvNeXt co-attention framework for
Vietnamese Visual Entailment.}
\label{fig:model_architecture}
\end{figure}

This section presents the PhoBERT--ConvNeXt framework for Vietnamese Visual
Entailment on OpenViVE. Given an image $I$ as the visual premise and a
Vietnamese hypothesis $H$, the model predicts one of three labels:
\textit{Entailment}, \textit{Contradiction}, or \textit{Neutral}. As
illustrated in Figure~\ref{fig:model_architecture}, the framework follows a
dual-stream design consisting of a ConvNeXt-based visual encoder, a
PhoBERT-based text encoder, a shared projection space, a single co-attention
fusion block, and a three-way classification head.

The cross-modal fusion module is adapted from the co-attention strategy
studied in METER~\cite{dou2022meter}, rather than being introduced as a new
attention mechanism. In our framework, a frozen ConvNeXt-Large visual
backbone~\cite{liu2022convnet} extracts spatial visual representations,
while a fine-tuned PhoBERT encoder~\cite{nguyen2020phobert} produces
contextualized representations of the Vietnamese hypothesis. Both modalities
are projected into a shared 512-dimensional space and interact through a
single bidirectional co-attention block before pooling and classification.
This interaction is particularly relevant to Visual Entailment, where the
predicted relation may depend on fine-grained differences in quantity,
attributes, object identity, actions, or spatial relations.

\subsection{Image Encoder}
\label{subsec:image_encoder}
For the visual branch, we use a pretrained ConvNeXt-Large~
\cite{liu2022convnet} as the image encoder. The ConvNeXt backbone remains
frozen throughout training, while the projection layer and subsequent
multimodal components are optimized for the VE task. Rather than applying
global pooling to the backbone output, we retain its final spatial feature
map for subsequent cross-modal interaction.

Given an input image $I$, the visual encoder produces a spatial feature map:
\begin{equation}
G = f_{\mathrm{img}}(I),
\end{equation}
where $f_{\mathrm{img}}(\cdot)$ denotes the frozen ConvNeXt-Large encoder.
For ConvNeXt-Large, the channel dimension of the resulting feature map is
$d_v=1536$. The spatial dimensions are flattened to form a sequence of
visual tokens:
\begin{equation}
F = [f_1,f_2,\ldots,f_M]
\in \mathbb{R}^{M\times d_v},
\qquad M=H\times W,
\end{equation}
where $H$ and $W$ denote the spatial dimensions of the feature map.

The visual tokens are normalized and linearly projected into the shared
$d$-dimensional multimodal space:
\begin{equation}
\bar{V}
=
\operatorname{LN}(F)W_v+b_v,
\end{equation}
where $W_v\in\mathbb{R}^{d_v\times d}$ and $d=512$.

To preserve spatial position information after flattening, a learnable
two-dimensional positional embedding $P_v$ is added to the projected visual
tokens:
\begin{equation}
V^{0}=\bar{V}+P_v,
\end{equation}
where $P_v\in\mathbb{R}^{M\times d}$. The positional embedding is initialized
on an $8\times8$ grid and resized using bicubic interpolation when the spatial
resolution of the feature map differs from the initialization size.

\subsection{Text Encoder}
\label{subsec:text_encoder}

For the textual branch, we use PhoBERT~\cite{nguyen2020phobert} to obtain
contextualized representations of the Vietnamese hypothesis. Unlike the
frozen visual backbone, the PhoBERT encoder is fine-tuned jointly with the
downstream Visual Entailment model.

Given a tokenized hypothesis $H$, the text encoder produces contextualized
token representations:
\begin{equation}
U = f_{\mathrm{text}}(H),
\end{equation}
where $U\in\mathbb{R}^{L\times d_t}$, $L$ denotes the token sequence length,
and $d_t$ is the hidden dimension of the selected PhoBERT variant.

The textual representations are then linearly projected into the same
shared hidden space as the visual tokens and subsequently normalized:
\begin{equation}
T^{0}
=
\operatorname{LN}(UW_t+b_t),
\end{equation}
where $W_t\in\mathbb{R}^{d_t\times d}$,
$T^{0}\in\mathbb{R}^{L\times d}$, and $d=512$.

\subsection{Co-Attention Fusion Block}
\label{subsec:co_attention}

To model cross-modal interaction, we employ a single co-attention block
inspired by the fusion strategy studied in METER~\cite{dou2022meter}.
The block is built on multi-head attention~\cite{vaswani2017attention}
and contains three stages: modality-specific self-attention, sequential
bidirectional cross-attention, and modality-specific feed-forward
transformations. The number of attention heads is treated as a
hyperparameter, while the hidden dimension is fixed at $d=512$.

First, self-attention is applied independently to the visual and textual
streams:
\begin{equation}
\tilde{V} =
\operatorname{LN}
\left(
V^{0} +
\operatorname{Dropout}
\left(
\operatorname{MHA}(V^{0},V^{0},V^{0})
\right)
\right),
\end{equation}

\begin{equation}
\tilde{T} =
\operatorname{LN}
\left(
T^{0} +
\operatorname{Dropout}
\left(
\operatorname{MHA}(T^{0},T^{0},T^{0};m)
\right)
\right),
\end{equation}
where $m$ denotes the textual padding mask. Padding positions in the textual
stream are masked during attention and excluded from subsequent textual
representations.

Next, cross-modal interaction is performed sequentially in both directions.
Visual tokens first attend to the textual representations:
\begin{equation}
V^{c} =
\operatorname{LN}
\left(
\tilde{V} +
\operatorname{Dropout}
\left(
\operatorname{MHA}
(Q=\tilde{V},K=\tilde{T},V=\tilde{T};m)
\right)
\right).
\end{equation}

The textual stream then attends to the updated visual representation:
\begin{equation}
T^{c} =
\operatorname{LN}
\left(
\tilde{T} +
\operatorname{Dropout}
\left(
\operatorname{MHA}
(Q=\tilde{T},K=V^{c},V=V^{c})
\right)
\right).
\end{equation}

This sequential bidirectional interaction produces a text-conditioned visual
representation and a visually conditioned textual representation before
pooling. Such interaction enables the two streams to exchange information
relevant to fine-grained VE decisions involving quantities, attributes,
object identities, actions, and spatial relations.

Finally, each modality is passed through an independent feed-forward network
with residual connection and layer normalization:
\begin{equation}
V^{1} =
\operatorname{LN}
\left(
V^{c} +
\operatorname{Dropout}
\left(
\operatorname{FFN}_{v}(V^{c})
\right)
\right),
\end{equation}

\begin{equation}
T^{1} =
\operatorname{LN}
\left(
T^{c} +
\operatorname{Dropout}
\left(
\operatorname{FFN}_{t}(T^{c})
\right)
\right).
\end{equation}

Each feed-forward network is defined as:
\begin{equation}
\operatorname{FFN}(x)
=
W_{2}\,
\operatorname{GELU}
\left(
W_{1}x+b_{1}
\right)
+b_{2},
\end{equation}
where the intermediate dimension is set to $2d$. With $d=512$, the
feed-forward hidden dimension is therefore 1024.

\subsection{Pooling and Multimodal Representation}
\label{subsec:pooling_fusion}

After the co-attention fusion block, the final visual and textual
representations are denoted by $V^{1}$ and $T^{1}$, respectively. The visual
tokens are aggregated using mean pooling:
\begin{equation}
z_v =
\frac{1}{M}
\sum_{i=1}^{M} V^{1}_{i},
\end{equation}
where $z_v \in \mathbb{R}^{d}$.

For the textual stream, masked mean pooling is applied so that padding
positions do not contribute to the aggregated representation:
\begin{equation}
z_t =
\frac{\sum_{j=1}^{L} m_j T^{1}_{j}}
     {\sum_{j=1}^{L} m_j},
\end{equation}
where $m_j \in \{0,1\}$ denotes the attention mask for the $j$-th token and
$z_t \in \mathbb{R}^{d}$.

The pooled visual and textual representations are concatenated to form the
final multimodal representation:
\begin{equation}
z =
\operatorname{Concat}(z_v,z_t)
\in \mathbb{R}^{2d}.
\end{equation}
With $d=512$, the resulting multimodal representation has a dimension of
1024.

\subsection{Classification Layer}
\label{subsec:classification_layer}

The multimodal representation $z$ is first regularized using dropout and
then passed to a linear classifier:
\begin{equation}
\tilde{z} = \operatorname{Dropout}(z),
\end{equation}
\begin{equation}
o = W_c \tilde{z} + b_c,
\end{equation}
where
$W_c \in \mathbb{R}^{3 \times 2d}$,
$b_c \in \mathbb{R}^{3}$, and
$o \in \mathbb{R}^{3}$ denotes the output logits for the three VE labels.

The corresponding class probabilities are obtained using softmax:
\begin{equation}
p = \operatorname{softmax}(o).
\end{equation}
The predicted label is then given by:
\begin{equation}
\hat{y}
=
\arg\max_{c \in \mathcal{Y}} p_c,
\qquad
\mathcal{Y}=\{E,C,N\}.
\end{equation}

\subsection{Training Objective}
\label{subsec:training_objective}

The model is optimized using the standard cross-entropy objective. Let
$\mathcal{D}$ denote the training set and
$\mathcal{Y}=\{E,C,N\}$ the set of VE labels. For each training sample,
the classifier produces logits $o_i$, from which the class probabilities
$p_i$ are obtained through softmax. The training objective is:

\begin{equation}
\mathcal{L}_{\mathrm{CE}}
=
-\frac{1}{\lvert\mathcal{D}\rvert}
\sum_{i=1}^{\lvert\mathcal{D}\rvert}
\sum_{c\in\mathcal{Y}}
y_{i,c}\log p_{i,c},
\label{eq:ce_loss}
\end{equation}

where $y_{i,c}$ is the one-hot ground-truth indicator and $p_{i,c}$ is the
predicted probability for class $c$.

Because OpenViVE contains an equal number of Entailment, Contradiction, and
Neutral examples, no class-specific weighting is applied. During training,
the ConvNeXt backbone remains frozen, while the visual projection layer,
learnable visual positional embeddings, PhoBERT encoder, text projection
layer, co-attention block, and classification head are optimized jointly.

\section{Experiments and Results}
\label{sec:experiments}

In this section, we evaluate different vision--language models on OpenViVE. The experiments are designed to examine how well supervised vision--language models, multimodal large language models, and our PhoBERT--ConvNeXt co-attention framework perform on Vietnamese Visual Entailment. All models are evaluated on the same development and test sets using Accuracy and macro F1-score.

\subsection{Experimental Setup}
\label{subsec:experimental_setup}

\subsubsection{Dataset and Evaluation Metrics}
\label{subsubsec:dataset_metrics}

All experiments are conducted on the predefined OpenViVE splits described in Section~\ref{subsubsec:initial_statistics}. Supervised models are trained on the training set, selected according to development performance, and finally evaluated on the test set. Since OpenViVE is balanced across the three labels, we report both Accuracy and macro F1-score.

Accuracy measures the proportion of correctly predicted samples:
\begin{equation}
\mathrm{Acc} = \frac{N_{\mathrm{correct}}}{N},
\end{equation}
where $N_{\mathrm{correct}}$ is the number of correctly classified samples and $N$ is the total number of evaluated samples.

For each class $c \in \{E,C,N\}$, precision, recall, and F1-score are computed as:
\begin{equation}
\mathrm{Precision}_{c} = \frac{TP_c}{TP_c + FP_c},
\end{equation}
\begin{equation}
\mathrm{Recall}_{c} = \frac{TP_c}{TP_c + FN_c},
\end{equation}
\begin{equation}
\mathrm{F1}_{c} = 
\frac{2 \cdot \mathrm{Precision}_{c} \cdot \mathrm{Recall}_{c}}
{\mathrm{Precision}_{c} + \mathrm{Recall}_{c}},
\end{equation}
where $TP_c$, $FP_c$, and $FN_c$ denote the true positives, false positives, and false negatives of class $c$, respectively.

We report macro F1-score by averaging the F1-score over the three labels:
\begin{equation}
\mathrm{F1}_{macro} = \frac{1}{3}\sum_{c \in \{E,C,N\}} \mathrm{F1}_{c}.
\end{equation}
Macro F1-score is important for Vietnamese Visual Entailment because the model should perform consistently across \textit{Entailment}, \textit{Contradiction}, and \textit{Neutral}, rather than relying on a single dominant label.

\subsubsection{Compared Models}
\label{subsubsec:compared_models}

To provide a comprehensive evaluation of OpenViVE, we compare four groups
of models, covering fine-tuned vision--language encoders, controlled
co-attention variants, zero-shot multimodal large language models (MLLMs),
and task-specific fine-tuned MLLMs.

\begin{itemize}

\item \textbf{Fine-tuned Vision--Language Baselines:}
We include TCL~\cite{yang2022tcl} and ALBEF~\cite{li2021albef} as
representative pretrained vision--language baselines. Both models are
fine-tuned on the OpenViVE training set and provide reference points for
supervised encoder-based performance on Vietnamese Visual Entailment.

\item \textbf{Co-attention Encoder Variants:}
We evaluate a family of models that share the same co-attention fusion
architecture described in Section~\ref{sec:framework} while varying the
visual and textual encoders. For the visual branch, we compare
Swin-v2~\cite{liu2022swinv2} and ConvNeXt~\cite{liu2022convnet}. For the
textual branch, we compare Vietnamese-specific PhoBERT~
\cite{nguyen2020phobert} with multilingual XLM-R~
\cite{conneau2020xlmr}, using both base and large variants where applicable.
This controlled comparison isolates the effects of the visual backbone,
text encoder, and language-specific representation on Vietnamese Visual
Entailment.

\item \textbf{Zero-shot Multimodal LLMs:}
We evaluate Qwen2.5-VL-7B, Qwen3-VL-8B, Gemma-3-4B-it, and
Ministral-3-8B without task-specific fine-tuning. Each model receives an
image together with a Vietnamese hypothesis and is instructed to predict
one of the three labels: \textit{Entailment}, \textit{Contradiction}, or
\textit{Neutral}. This setting evaluates the extent to which general-purpose
MLLMs transfer to Vietnamese Visual Entailment without exposure to OpenViVE
training examples. The zero-shot prompt template is provided in
Appendix~\ref{app:zeroshot_prompt}.

\item \textbf{Fine-tuned Multimodal LLMs:}
Finally, we fine-tune Qwen3-VL-2B and Qwen3.5-VL-0.8B on the OpenViVE
training set. These experiments examine how task-specific adaptation affects
the performance of generative multimodal models and provide a direct
comparison with the specialized encoder-based architectures evaluated above.

\end{itemize}

\subsubsection{Implementation Details}
\label{subsubsec:implementation_details}

All experiments are implemented in PyTorch and conducted on an NVIDIA L4 GPU. To ensure controlled comparisons among the encoder-based variants, we keep the classification head, loss function, preprocessing pipeline, optimizer, and training protocol fixed, while varying only the visual encoder, text encoder, or fusion configuration according to each experiment.

For image preprocessing, each input image is resized to $232\times232$,
center-cropped to $224\times224$, converted to a tensor, and normalized
using ImageNet statistics. Vietnamese hypotheses are tokenized using the
tokenizer associated with the corresponding language encoder and padded or
truncated to a maximum sequence length of 256 tokens. The training batch
size is set to 32.

For the PhoBERT--ConvNeXt framework, the pretrained ConvNeXt visual
backbone remains frozen throughout training, whereas the PhoBERT encoder and
all downstream multimodal components are optimized for the VE task. Visual
and textual representations are projected into a shared 512-dimensional
space. Unless otherwise specified, the default model uses a single
co-attention block with eight attention heads and a dropout rate of 0.1.
The ablation study separately evaluates the removal of the co-attention
block and different numbers of attention heads.

All supervised encoder-based models are trained using the cross-entropy
objective and optimized with AdamW using a learning rate of
$3\times10^{-5}$ and a weight decay of 0.01. Training is conducted for up
to five epochs with a linear learning-rate schedule and 10\% warm-up.
Gradient clipping with a maximum norm of 1.0 is applied during training.
The best checkpoint is selected according to performance on the development
set and subsequently evaluated on the test set.

For zero-shot multimodal LLMs, no parameter updates are performed. Each
model receives an image, a Vietnamese hypothesis, and the task instruction,
and is required to return one of the three VE labels. The zero-shot prompt
template is provided in Appendix~\ref{app:zeroshot_prompt}.For fine-tuning the multimodal LLMs, we employ LoRA with rank $r = 32$ and scaling factor $\alpha = 32$, applied across both attention and MLP modules in the visual and language components. The models are optimized using AdamW with an initial learning rate of $2 \times 10^{-4}$ and trained for 3 epochs.

\subsection{Main Results}
\label{subsec:main_results}

Table~\ref{tab:main_results} reports Accuracy and Macro-F1 on the OpenViVE
development and test sets. The evaluated systems are grouped into
fine-tuned vision--language baselines, co-attention encoder variants,
zero-shot multimodal LLMs, and fine-tuned multimodal LLMs. For the
co-attention encoder variants, we report the default configuration using a
single co-attention block with $h=8$ attention heads. The effect of removing
the co-attention block and varying the number of attention heads is examined
separately in Section~\ref{subsec:ablation_study}.

\begin{table*}[!htbp]
\centering
\normalsize
\renewcommand{\arraystretch}{1.22}
\setlength{\tabcolsep}{8pt}
\caption{Experimental results on the OpenViVE development and test sets.}
\label{tab:main_results}
\begin{tabular*}{0.95\textwidth}{@{\extracolsep{\fill}}lrrrr@{}}
\toprule
\textbf{Model} 
& \multicolumn{2}{c}{\textbf{Dev}} 
& \multicolumn{2}{c}{\textbf{Test}} \\
\cmidrule(lr){2-3} \cmidrule(lr){4-5}
& \textbf{Acc.} & \textbf{F1} & \textbf{Acc.} & \textbf{F1} \\
\midrule

\multicolumn{5}{c}{\textbf{Fine-tuned Vision--Language Baselines}} \\
\midrule
TCL~\cite{yang2022tcl}
& 64.84 & 64.92 & 65.24 & 65.26 \\
ALBEF~\cite{li2021albef}
& 63.95 & 63.83 & 64.40 & 64.22 \\

\midrule
\multicolumn{5}{c}{\textbf{Co-attention Encoder Variants ($h=8$)}} \\
\midrule
Swin-v2 + XLM-R$_{\text{base}}$
& 82.27 & 82.27 & 82.16 & 82.18 \\
Swin-v2 + XLM-R$_{\text{large}}$
& 82.82 & 82.92 & 82.32 & 82.44 \\
Swin-v2 + PhoBERT$_{\text{base}}$
& 84.15 & 84.04 & 83.88 & 83.74 \\
Swin-v2 + PhoBERT$_{\text{large}}$
& 84.46 & 84.30 & 84.22 & 84.09 \\
ConvNeXt + XLM-R$_{\text{base}}$
& 82.82 & 82.82 & 81.97 & 81.98 \\
ConvNeXt + XLM-R$_{\text{large}}$
& 83.06 & 83.11 & 82.84 & 82.90 \\
ConvNeXt + PhoBERT$_{\text{base}}$
& 84.57 & 84.61 & 83.91 & 83.96 \\
\textbf{ConvNeXt + PhoBERT$_{\text{large}}$}
& 84.67 & 84.69 & \textbf{84.84} & \textbf{84.90} \\

\midrule
\multicolumn{5}{c}{\textbf{Zero-shot Multimodal LLMs}} \\
\midrule
Qwen2.5-VL-7B
& 50.02 & 48.87 & 49.31 & 46.39 \\
Qwen3-VL-8B
& 56.11 & 53.72 & 54.22 & 51.93 \\
Gemma-3-4B-it
& 33.40 & 22.10 & 33.54 & 20.49 \\
Ministral-3-8B
& 44.03 & 40.02 & 41.97 & 38.14 \\

\midrule
\multicolumn{5}{c}{\textbf{Fine-tuned Multimodal LLMs}} \\
\midrule
Qwen3-VL-2B
& 81.01 & 81.02 & 80.78 & 80.77 \\
Qwen3.5-VL-0.8B
& \textbf{85.23} & \textbf{85.26} & 84.75 & 84.79 \\

\bottomrule
\end{tabular*}
\end{table*}

As shown in Table~\ref{tab:main_results}, ConvNeXt +
PhoBERT$_{\text{large}}$ achieves the highest observed test scores among
the evaluated models, reaching 84.84\% Accuracy and 84.90\% Macro-F1.
On the development set, fine-tuned Qwen3.5-VL-0.8B obtains the highest
scores, with 85.23\% Accuracy and 85.26\% Macro-F1. On the test set, the
two models perform very similarly: ConvNeXt +
PhoBERT$_{\text{large}}$ exceeds Qwen3.5-VL-0.8B by only 0.09 percentage
points in Accuracy and 0.11 percentage points in Macro-F1.

The co-attention encoder variants consistently outperform the fine-tuned
TCL and ALBEF baselines, with test Macro-F1 scores ranging from 81.98\%
to 84.90\%, compared with 65.26\% and 64.22\% for TCL and ALBEF,
respectively. Within matched co-attention configurations, PhoBERT also
consistently outperforms the corresponding XLM-R variant. On the test set,
the Macro-F1 gains obtained by replacing XLM-R with PhoBERT are 1.56 and
1.65 percentage points for the Swin-v2 base and large configurations, and
1.98 and 2.00 percentage points for the corresponding ConvNeXt
configurations. These comparisons provide evidence that Vietnamese-specific
text representations are beneficial within the evaluated co-attention
architecture.

Zero-shot multimodal LLMs perform substantially below the supervised models
in this evaluation. Among the zero-shot models, Qwen3-VL-8B achieves the
highest test result, with 54.22\% Accuracy and 51.93\% Macro-F1. In
contrast, the task-adapted Qwen3-VL-2B and Qwen3.5-VL-0.8B obtain
considerably higher scores, with Qwen3.5-VL-0.8B reaching 84.75\%
Accuracy and 84.79\% Macro-F1. Although the zero-shot and fine-tuned MLLM
groups are not directly size-matched, the results reveal a substantial gap
between general zero-shot transfer and task-specific supervised adaptation
on OpenViVE. At the same time, the specialized ConvNeXt--PhoBERT framework
remains highly competitive with the fine-tuned MLLMs and achieves the
highest observed test scores in the comparison.

\subsection{Ablation Study}
\label{subsec:ablation_study}

To examine the contribution of cross-modal fusion, we conduct an ablation
study on the PhoBERT--ConvNeXt framework. We compare models without
co-attention against models using a single co-attention block with different
numbers of attention heads. In the no-co-attention setting, the visual and
textual representations are encoded independently, pooled, and concatenated
directly before classification. In the co-attention setting, the two
modalities interact through the bidirectional fusion block described in
Section~\ref{subsec:co_attention} before pooling and prediction. This
comparison evaluates the contribution of explicit image--text interaction
and the sensitivity of the framework to the number of attention heads.

\begin{table*}[!htbp]
\centering
\small
\renewcommand{\arraystretch}{1.18}
\setlength{\tabcolsep}{4pt}

\caption{Ablation results for co-attention fusion and attention-head
configurations in the PhoBERT--ConvNeXt framework. Accuracy and F1 are
reported in percentage, where F1 denotes Macro-F1.}
\label{tab:ablation_coattention}

\resizebox{0.98\textwidth}{!}{%
\begin{tabular}{llcrrrrr}
\toprule

\textbf{Model}
& \textbf{Fusion}
& \textbf{Heads}
& \textbf{Params (M)}
& \multicolumn{2}{c}{\textbf{Dev}}
& \multicolumn{2}{c}{\textbf{Test}} \\

\cmidrule(lr){5-6}
\cmidrule(lr){7-8}

& & & &
\textbf{Acc.} & \textbf{F1}
& \textbf{Acc.} & \textbf{F1} \\

\midrule

\multirow{4}{*}{
\shortstack[l]{PhoBERT$_{\text{base}}$\\+ ConvNeXt}}
& No Co-Attn. & -- & 332.4
& 73.78 & 74.23 & 73.51 & 73.99 \\

& Co-Attn. & 4 & 338.7
& 84.26 & 84.29 & \textbf{84.34} & \textbf{84.39} \\

& Co-Attn. & 8 & 338.7
& \textbf{84.57} & \textbf{84.61} & 83.91 & 83.96 \\

& Co-Attn. & 16 & 338.7
& 84.14 & 83.88 & 84.16 & 83.91 \\

\midrule

\multirow{4}{*}{
\shortstack[l]{PhoBERT$_{\text{large}}$\\+ ConvNeXt}}
& No Co-Attn. & -- & 566.7
& 74.89 & 75.27 & 74.12 & 75.12 \\

& Co-Attn. & 4 & 573.0
& \textbf{85.16} & \textbf{85.19} & 84.04 & 84.09 \\

& Co-Attn. & 8 & 573.0
& 84.67 & 84.69 & \textbf{84.84} & \textbf{84.90} \\

& Co-Attn. & 16 & 573.0
& 83.77 & 83.80 & 84.14 & 84.19 \\

\bottomrule
\end{tabular}%
}
\end{table*}

As shown in Table~\ref{tab:ablation_coattention}, removing the co-attention
block results in a substantial decrease in performance for both PhoBERT
variants. For PhoBERT$_{\text{base}}$ + ConvNeXt, the model without
co-attention achieves 73.99\% test Macro-F1, whereas the best co-attention
configuration reaches 84.39\%, corresponding to an improvement of
10.40 percentage points. A similar pattern is observed for
PhoBERT$_{\text{large}}$ + ConvNeXt, where test Macro-F1 increases from
75.12\% without co-attention to 84.90\% with the $h=8$ configuration, a
gain of 9.78 percentage points. These gains are observed despite an increase
of only 6.3M parameters when the co-attention block is introduced. The
results therefore provide strong evidence that explicit cross-modal
interaction contributes substantially to performance within the proposed
framework.

The number of attention heads does not exhibit a monotonic relationship with
performance. For PhoBERT$_{\text{base}}$, the $h=8$ configuration achieves
the highest development Macro-F1, whereas $h=4$ obtains the highest test
Macro-F1. For PhoBERT$_{\text{large}}$, $h=4$ performs best on the
development set, while $h=8$ achieves the highest test performance.
Increasing the number of heads to $h=16$ does not improve either variant
consistently. These results suggest that increasing attention granularity
alone does not guarantee better cross-modal representations, and that
performance is relatively sensitive to the selected head configuration.

Overall, the ablation study shows that the principal architectural difference
between the two settings---the presence or absence of co-attention---is
associated with a large performance gap, whereas changing the number of
attention heads produces comparatively smaller variations. We therefore use
$h=8$ as the default configuration for the co-attention encoder variants in
the main experiments.

\subsection{Result Analysis}
\label{subsec:result_analysis}

Beyond the aggregate results reported in Table~\ref{tab:main_results}, we
further analyze model behavior from four perspectives. We examine the effect
of hypothesis length, qualitatively inspect cross-modal attention patterns,
analyze class-level error distributions through the confusion matrix, and
evaluate residual textual artifacts using a hypothesis-only setting. These
analyses provide complementary insights into the conditions under which the
models succeed or fail on OpenViVE.

\subsubsection{Effect of Hypothesis Length}
\label{subsubsec:length_effect}

\begin{figure}[!htbp]
\centering
\includegraphics[width=0.95\columnwidth]{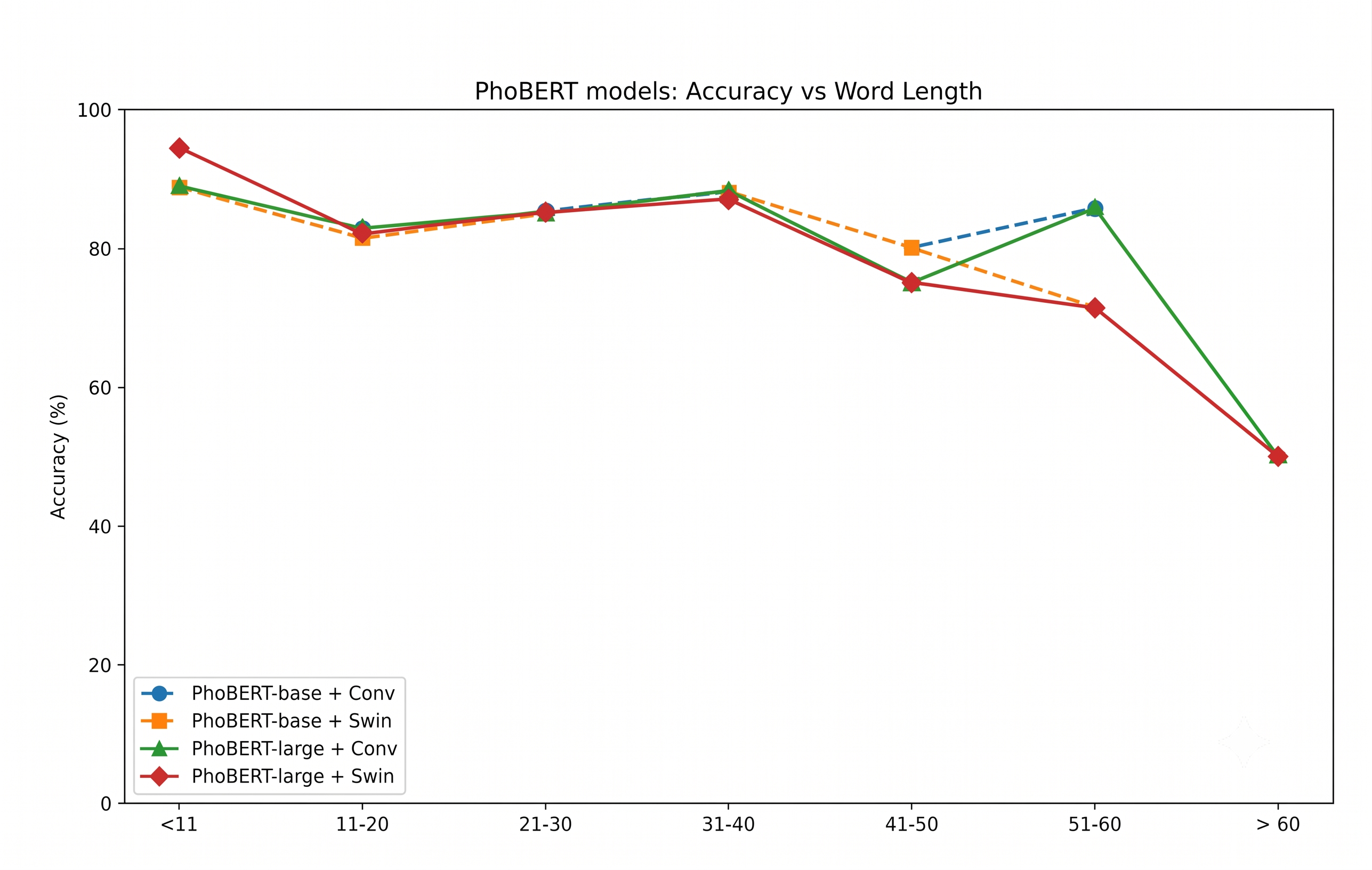}
\caption{Accuracy of PhoBERT--ConvNeXt and PhoBERT--Swin-v2 co-attention
variants across different hypothesis-length intervals.}
\label{fig:length_effect}
\end{figure}

We further examine whether model performance varies with hypothesis length.
Figure~\ref{fig:length_effect} reports the accuracy of four co-attention
variants combining PhoBERT$_{\text{base}}$ or
PhoBERT$_{\text{large}}$ with either ConvNeXt or Swin-v2 across different
hypothesis-length intervals. For hypotheses containing up to 40 words, all
four configurations exhibit relatively stable performance, with accuracy
generally remaining above 80\%. This pattern is observed across both visual
backbones and both PhoBERT variants.

The models become more variable for hypotheses longer than 40 words.
In the 41--50 word interval, the ConvNeXt and Swin-v2 configurations show
noticeable differences, while a similar divergence appears in the 51--60
word interval. In particular, the ConvNeXt-based variants recover to higher
accuracy in the 51--60 range, whereas the Swin-v2-based variants remain
lower. These fluctuations suggest that performance on longer hypotheses may
depend not only on textual complexity but also on the interaction between
the language and visual encoders.

Accuracy drops sharply to approximately 50\% for hypotheses exceeding 60 words across all configurations. This degradation is driven by the accumulation of multiple entities, modifiers, and contextual clauses, which substantially compounds the difficulty of fine-grained cross-modal verification. While very long hypotheses constitute a lower-density segment of the dataset, this consistent drop highlights compositional reasoning as a major remaining challenge on OpenViVE

\subsubsection{Attention Visualization}
\label{subsubsec:attention_visualization}

\begin{figure*}[!t]
\centering
\footnotesize
\renewcommand{\arraystretch}{1.18}
\setlength{\tabcolsep}{2.5pt}

\begin{tabular}{
    >{\centering\arraybackslash}m{0.17\textwidth}
    >{\centering\arraybackslash}m{0.13\textwidth}
    >{\centering\arraybackslash}m{0.20\textwidth}
    >{\raggedright\arraybackslash}m{0.42\textwidth}
}
\toprule
\textbf{Original Image}
& \textbf{Label}
& \textbf{Attention Map}
& \textbf{Vietnamese Hypothesis}
\\
\midrule

&
\textit{Entailment}
& \includegraphics[width=0.19\textwidth]{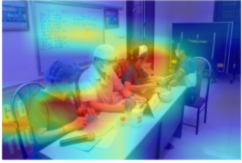}
& \textit{Có hai người nữ mặc đồng phục màu xanh và một người nam mặc áo trắng đang ngồi bên một chiếc bàn dài.}
(\textit{There are two women in green uniforms and one man in a white shirt sitting at a long table}).
\\[0.8em]

\includegraphics[width=0.15\textwidth]{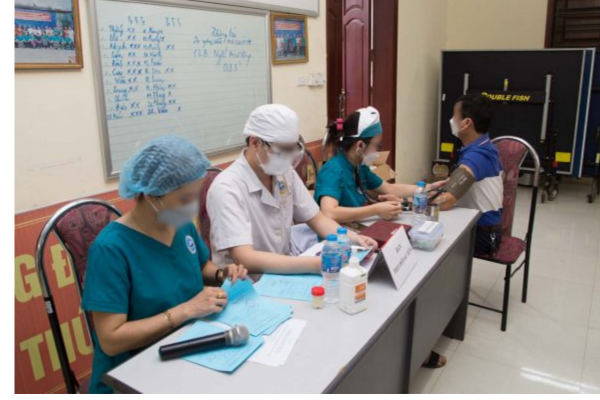}
&
\textit{Contradiction}
& \includegraphics[width=0.19\textwidth]{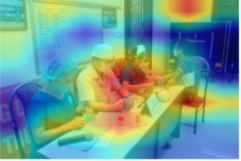}
& \textit{Chỉ có một người nữ mặc đồng phục màu xanh và một người nam mặc áo trắng đang ngồi bên một chiếc bàn dài.}
(\textit{There is only one woman in a green uniform and one man in a white shirt sitting at a long table}).
\\[0.8em]

&
\textit{Neutral}
& \includegraphics[width=0.19\textwidth]{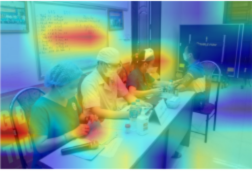}
& \textit{Các bác sĩ đang sử dụng thiết bị y tế tiên tiến nhất để đảm bảo rằng tất cả các bệnh nhân đều nhận được dịch vụ chăm sóc tốt nhất có thể trong quá trình kiểm tra sức khỏe.}
(\textit{The doctors are using the most advanced medical equipment to ensure that all patients receive the best possible care during the health check-up}).
\\

\bottomrule
\end{tabular}

\caption{Attention visualization for an example in which the model correctly predicts all three hypotheses associated with the same image.}
\label{fig:attention_sample2}
\end{figure*}

\begin{figure*}[!t]
\centering
\footnotesize
\renewcommand{\arraystretch}{1.18}
\setlength{\tabcolsep}{2.5pt}

\begin{tabular}{
    >{\centering\arraybackslash}m{0.17\textwidth}
    >{\centering\arraybackslash}m{0.13\textwidth}
    >{\centering\arraybackslash}m{0.20\textwidth}
    >{\raggedright\arraybackslash}m{0.42\textwidth}
}
\toprule
\textbf{Original Image}
& \textbf{Label}
& \textbf{Attention Map}
& \textbf{Vietnamese Hypothesis}
\\
\midrule

&
\textit{Entailment}
& \includegraphics[width=0.19\textwidth]{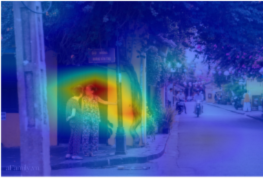}
& \textit{Trên đường Hoàng Văn Thụ, hai cụ bà mặc quần áo hoa đứng tâm sự; phía sau là những ngôi nhà sơn vàng cổ kính và xe máy đang di chuyển.}
(\textit{On Hoang Van Thu Street, two elderly women in floral clothes are standing and talking; behind them are old yellow-painted houses and moving motorbikes}).
\\[0.8em]

\includegraphics[width=0.15\textwidth]{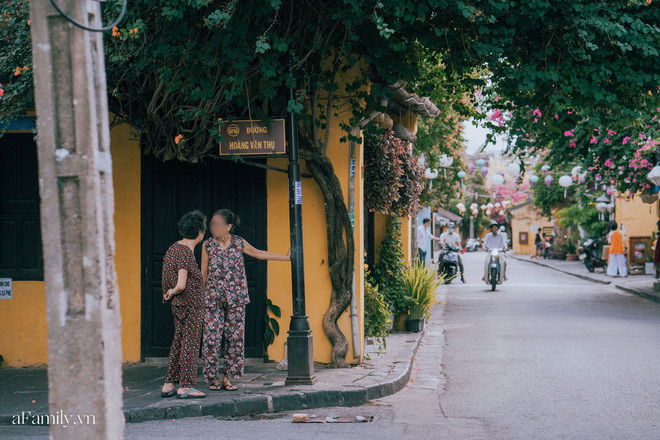}
&
\textit{Contradiction}
& \includegraphics[width=0.19\textwidth]{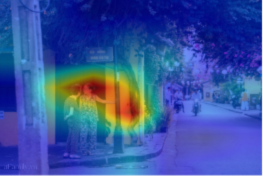}
&\textit{ Hai người đàn ông lớn tuổi mặc đồ bộ hoa đang trò chuyện thân mật bên cạnh cột đèn trên vỉa hè của một con đường nhỏ ở Hội An.}
(\textit{Two elderly men in floral outfits are having a close conversation next to a lamp post on the sidewalk of a small street in Hoi An}).
\\[0.8em]

&
\textit{Neutral}
& \includegraphics[width=0.19\textwidth]{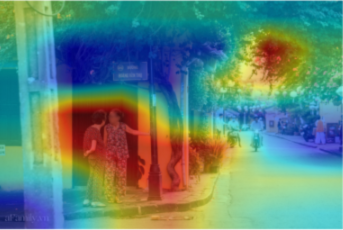}
& \textit{Hai bà lão mặc áo quần hoa, ôn lại chuyện thời trẻ ở Hội An, nhớ về những tháng ngày tươi đẹp trên con phố này.}
(\textit{Two elderly women in floral clothes are reminiscing about their youth in Hoi An and the beautiful days on this street}).
\\

\bottomrule
\end{tabular}

\caption{Attention visualization for a challenging example. The model correctly predicts the entailment and neutral hypotheses, but fails on the contradiction hypothesis.}
\label{fig:attention_sample1}
\end{figure*}

To qualitatively inspect the cross-modal attention patterns learned by the
PhoBERT--ConvNeXt framework, we visualize representative attention maps for
two examples in OpenViVE. Figures~\ref{fig:attention_sample2}
and~\ref{fig:attention_sample1} show the original image, the attention maps
associated with different hypotheses, and the corresponding Vietnamese and
English hypothesis sentences.

Figure~\ref{fig:attention_sample2} presents an example in which the model correctly predicts all three labels. The attention maps show that the model consistently focuses on visually salient regions related to the hypothesis, especially the people sitting at the table and the surrounding contextual objects. For the entailment hypothesis, the attended regions correspond to the people described in the sentence. For the contradiction hypothesis, although the sentence changes the number of women, the attention remains concentrated on the relevant human region,and the model produces the correct contradiction prediction. For the neutral hypothesis, the attention map remains related to the medical scene, but the sentence introduces unverifiable information about the use of advanced medical equipment and service quality, which cannot be confirmed solely from the image.

Figure~\ref{fig:attention_sample1} illustrates a challenging case that reveals a limitation of the current model. The model correctly predicts the entailment and neutral hypotheses, but fails on the contradiction hypothesis. Here, the contradiction is not caused by an unrelated object or an obvious scene mismatch; instead, it depends on a fine-grained attribute change, where the visible elderly women are described as elderly men. Although the attention map still highlights the relevant human region, the model fails to verify the specific semantic attribute required for the final decision. This suggests that successful attention to the correct visual area does not necessarily imply successful visual-textual inference. For OpenViVE, many contradiction examples require precise verification of attributes, quantities, object identities, and visual appearances, making fine-grained contradiction detection a key remaining challenge.

Overall, the visualizations show that the model often assigns attention to image regions that are semantically related to the corresponding hypotheses. However, these attention maps should be interpreted as qualitative inspection of the model's attention patterns rather than as causal explanations of its predictions. The failure case further indicates
that attending to a relevant visual region is not sufficient for reliable
visual-textual inference when the predicted relation depends on a
fine-grained semantic distinction.

\subsubsection{Error Analysis}
\label{subsubsec:error_analysis}

\begin{figure}[!htbp]
\centering
\includegraphics[width=0.65\columnwidth]{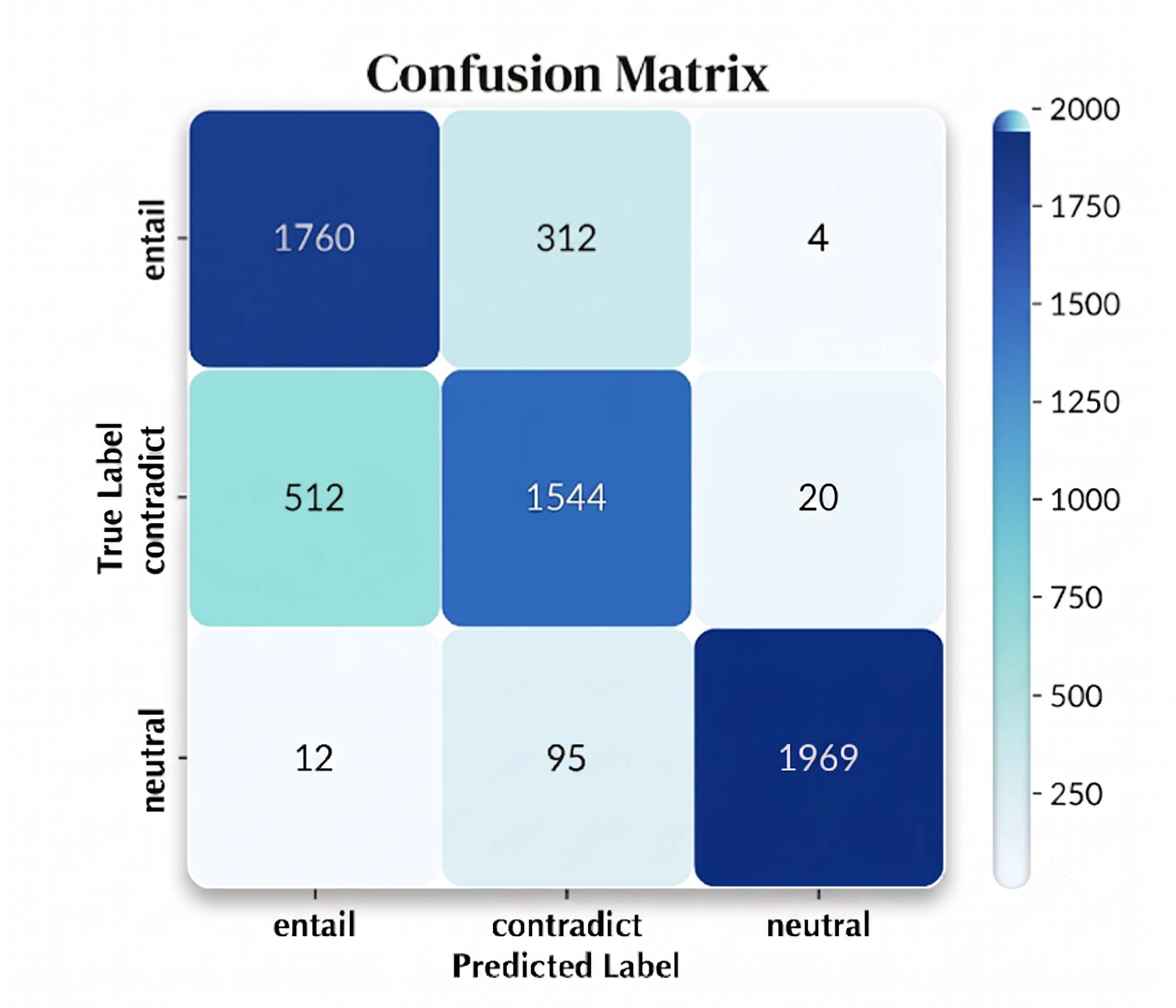}
\caption{Confusion matrix of the ConvNeXt +
PhoBERT$_{\text{large}}$ model ($h=8$) on the OpenViVE development set.}
\label{fig:confusion_matrix}
\end{figure}

Figure~\ref{fig:confusion_matrix} presents the confusion matrix of the
ConvNeXt + PhoBERT$_{\text{large}}$ model with $h=8$ on the OpenViVE
development set. Since OpenViVE is balanced across the three labels, the
matrix provides a direct view of the class-specific prediction patterns and
the main remaining sources of error.

The model performs best on the \textit{Neutral} class, correctly predicting
1,969 out of 2,076 examples, corresponding to a class-wise recall of
approximately 94.8\%. For \textit{Entailment}, 1,760 out of 2,076 examples
are correctly classified, resulting in a recall of approximately 84.8\%.
In contrast, \textit{Contradiction} is the most difficult class, with
1,544 correctly predicted examples and a recall of approximately 74.4\%.

Most errors occur between \textit{Entailment} and \textit{Contradiction}.
Specifically, 512 contradiction examples are misclassified as entailment,
whereas 312 entailment examples are predicted as contradiction. This
asymmetry indicates that the model has greater difficulty recognizing
contradictory evidence when the hypothesis remains broadly compatible with
the overall visual scene.

This error pattern reflects the fine-grained nature of many contradiction
examples in OpenViVE. Entailment and contradiction hypotheses may refer to
similar entities, actions, and visual contexts while differing in only a
single semantic detail, such as quantity, object identity, attribute, action,
or spatial relation. As also observed in the attention visualization, the
model may attend to a relevant visual region while still failing to verify
the specific semantic detail required to distinguish contradiction from
entailment.

Overall, contradiction recognition remains the main source of classification
error for the evaluated model. Future improvements may therefore benefit
from stronger mechanisms for fine-grained attribute comparison, counting,
relation verification, and phrase-level visual grounding.

\subsubsection{Hypothesis-only Artifact Analysis}
\label{subsubsec:hypothesis_only_artifact}

To further examine whether OpenViVE contains exploitable textual artifacts,
we conduct a hypothesis-only analysis using a PhoBERT-based text classifier.
In this setting, the classifier receives only the Vietnamese hypothesis without
access to the corresponding image. The same classifier architecture and
training protocol are used before and after artifact refinement so that the
performance difference reflects changes in hypothesis-only predictability
rather than changes in the model. High hypothesis-only performance would
indicate that label-specific lexical or stylistic patterns can be exploited to
predict the target label without using visual evidence.

\begin{figure}[!htbp]
\centering
\includegraphics[width=0.85\columnwidth]{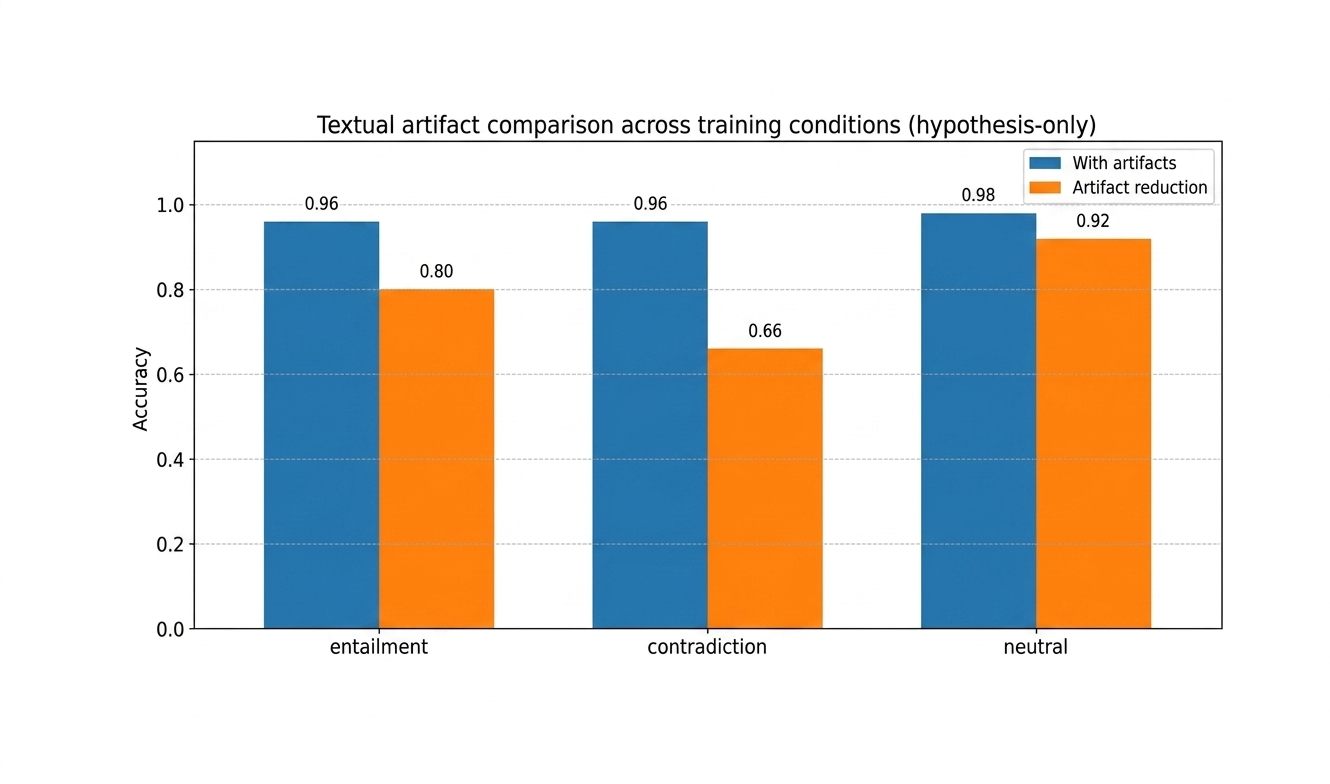}
\caption{Class-wise hypothesis-only accuracy before and after artifact
reduction.}
\label{fig:hypothesis_only_artifact}
\end{figure}

Figure~\ref{fig:hypothesis_only_artifact} compares the class-wise
hypothesis-only accuracy before and after artifact reduction. Before
refinement, the hypothesis-only model achieves very high accuracy for all
three labels, reaching 0.96 for \textit{Entailment}, 0.96 for
\textit{Contradiction}, and 0.98 for \textit{Neutral}. These results indicate
that early versions of the generated hypotheses contained strong
label-discriminative textual patterns that could be exploited without access
to the image.

After artifact reduction, hypothesis-only performance decreases
substantially for \textit{Entailment} and \textit{Contradiction}.
Contradiction accuracy drops from 0.96 to 0.66, while Entailment decreases
from 0.96 to 0.80. These reductions indicate that the rewriting and
refinement process weakens label-predictive lexical patterns and makes the
task less exploitable from the hypothesis alone.

The \textit{Neutral} class remains considerably more predictable from text,
with a hypothesis-only accuracy of 0.92 after refinement. This suggests that
residual linguistic cues remain, particularly in hypotheses containing
information about purpose, temporal context, intention, or background
events. Therefore, the artifact-reduction process should be interpreted as
reducing rather than eliminating textual shortcuts. Overall, the analysis
shows that the refinement procedure substantially reduces hypothesis-only
predictability for Entailment and Contradiction, while residual artifacts in
Neutral remain an important limitation of the current dataset.

\section{Conclusion and Future Work}\label{sec:conclusions}

In this paper, we addressed the challenges of vision-language semantic reasoning in low-resource language scenarios by investigating the Visual Entailment (VE) task for Vietnamese. To bridge the existing resource gap, we constructed and evaluated OpenViVE, a large-scale, well-structured, and label-balanced Vietnamese VE benchmark dataset comprising 62,268 image--hypothesis pairs. The dataset was developed through a hybrid pipeline leveraging Large Language Models (LLMs) for label-specific hypothesis generation alongside human quality control, which involved formal rule design, inter-annotator agreement metrics, and hypothesis-only artifact reduction. Analytical results showed that OpenViVE provides reliable Vietnamese hypotheses while reducing label-specific textual artifacts.

Furthermore, a core contribution of our work is the proposal of the PhoBERT--ConvNeXt framework, a two-stream co-attention architecture tailored for Vietnamese vision-language inference. By benchmarking various state-of-the-art vision-language model (VLM) architectures, zero-shot multimodal LLMs, and fine-tuned multimodal LLMs on OpenViVE, experimental results demonstrated that the PhoBERT--ConvNeXt framework consistently outperforms the corresponding multilingual counterparts (e.g., XLM-R) and the evaluated zero-shot configurations. Benefiting from the specialized Vietnamese textual encodings of PhoBERT and the visual feature representations of ConvNeXt, this integrated framework achieved the highest observed test Accuracy and Macro-F1 of 84.84\% and 84.90\%, respectively. In-depth behavioral evaluations via confusion matrices, hypothesis-length intervals, attention visualization, and label-wise Pointwise Mutual Information (PMI) shed light on model prediction traits. Notably, while the framework achieved its highest class-wise recall on the Neutral label, the primary bottleneck persisted within the fine-grained semantic boundary separating the Entailment and Contradiction classes. Ultimately, OpenViVE and the PhoBERT--ConvNeXt framework provide an evaluation benchmark and strong baseline for future research in Vietnamese vision-language inference.

Despite the notable contributions of OpenViVE, several open challenges remain to be addressed in future studies. We outline three key directions to extend this line of research: (1) Although Pointwise Mutual Information (PMI) analysis and constrained prompting were used to reduce hypothesis-only biases, text-based annotation artifacts cannot be completely eliminated when utilizing LLMs. Future work will investigate more sophisticated adversarial filtering techniques and contrastive learning objectives to reduce the model's reliance on shortcuts or textual signals without cross-referencing visual premises. (2) The current version of OpenViVE primarily focuses on object descriptions, spatial relations, and straightforward actions. To push the boundaries of multimodal reasoning, we plan to extend the dataset by incorporating more complex, high-level reasoning types, such as spatial relational reasoning (e.g., similar to NLVR2~\cite{suhr2019nlvr2}), numerical quantification, and cultural commonsense reasoning deeply rooted in Vietnamese contexts. (3) Given that our strongest encoder-based model relies on a discriminative two-stream co-attention paradigm, a promising next step is to broaden the evaluation of generative VLMs, such as LLaVA~\cite{liu2023llava}, BLIP-2~\cite{li2023blip2}, and native multilingual foundation models, beyond the models considered in this study. We aim to investigate parameter-efficient fine-tuning (PEFT) methods, such as LoRA~\cite{hu2022lora} or prompt-tuning~\cite{lester2021power}, to adapt these large generative models for structured multi-class cross-modal reasoning without requiring excessive computational overhead.

\section*{Acknowledgement}
This research is funded by Vietnam National University HoChiMinh City (VNU-HCM) under grant number NCM2025-26-02.

\section*{Funding Declaration}
This research is funded by Vietnam National University HoChiMinh City (VNU-HCM) under grant number NCM2025-26-02.

\section*{Data Availability}

Data will be made available on reasonable request.

\section*{CRediT authorship contribution statement}
 \textbf{Bao Quoc Trinh:} Conceptualization; Data curation; Formal analysis; Investigation; Methodology; Validation; Visualization; Writing - original draft. 
 
 \textbf{Tin Van Huynh:} Conceptualization; Data curation; Formal analysis; Investigation; Validation; Supervision; Writing - review \& editing. 
 
 \textbf{Kiet Van Nguyen:} Conceptualization; Formal analysis; Investigation; Methodology; Validation; Supervision; Writing - review \& editing. 

  \textbf{Ngan Luu-Thuy Nguyen:} Conceptualization; Formal analysis; Investigation; Methodology; Validation; Supervision; Writing - review \& editing.

\section*{Declaration of Competing Interest}
The authors declare that they have no known competing financial interests or personal relationships that could have appeared to influence the work reported in this article.

\bibliography{sn-bibliography}

\appendix

\section{Zero-shot Evaluation Prompt} \label{app:zeroshot_prompt} For zero-shot evaluation of multimodal LLMs, we use the following instruction template. The input image is provided together with a Vietnamese hypothesis, and the model is required to return exactly one of the three predefined labels. 

\begin{promptcard}{Zero-shot Prompt for Vietnamese Visual Entailment} 

\promptsection{Role} You are an expert in the Visual Entailment task. Task. You are given an image and a hypothesis sentence written in Vietnamese. Your goal is to determine the semantic relationship between the image and the hypothesis. 

\promptsection{Label definitions} 
\begin{itemize}[leftmargin=*, itemsep=2pt, topsep=2pt] 
\item Entailment: The hypothesis is clearly supported by the image. 
\item Contradiction: The hypothesis is clearly refuted by the image. 
\item Neutral: The hypothesis cannot be verified or refuted based solely on the image. 
\end{itemize} 
\promptsection{Input}
\begin{itemize}[leftmargin=*, itemsep=2pt, topsep=2pt] 
\item Image: provided separately. 
\item Vietnamese hypothesis: ``[HYPOTHESIS]''
\end{itemize} 
\promptsection{Output} 

Return exactly one label: Entailment, Contradiction, or Neutral. \end{promptcard}

\section{Prompt Templates for Hypothesis Generation} \label{app:generation_prompts} 

This appendix presents the label-specific prompt templates used for generating Vietnamese hypotheses in OpenViVE. The original prompts were written in Vietnamese during dataset construction, and all generated hypotheses were produced in Vietnamese. For readability, we provide an English translation of the templates while preserving the original task instructions, label definitions, and generation constraints used in the dataset construction pipeline. 
\subsection{Entailment Generation Prompt} \label{app:entailment_prompt} 

\begin{promptcard}{ENTAILMENT Prompt for Visual Entailment} 
\promptsection{Role} You are an expert in vision--language data construction for Vietnamese Visual Entailment. Your task is to generate high-quality Vietnamese hypotheses for the Entailment label. 

\promptsection{Label definition} A hypothesis is labeled as Entailment if the image provides sufficient visual evidence to conclude that the hypothesis is true. 

\promptsection{Task} Given an image, generate two different Vietnamese hypotheses that are both entailed by the image. 

\promptsection{Input} \begin{itemize}[leftmargin=*, itemsep=1pt, topsep=2pt] \item Image: provided separately. \end{itemize}

\promptsection{Generation requirements} \begin{enumerate}[leftmargin=*, itemsep=2pt, topsep=2pt] \item The hypotheses must be directly grounded in visible evidence from the image. Do not introduce objects, events, attributes, or background facts that cannot be observed or reasonably inferred. \item The hypotheses may include simple logical inferences from visible actions or scene evidence. For example, if a person is holding an umbrella and the ground is wet, the hypothesis may describe the person as moving in rainy or post-rain conditions. \item The hypotheses must be natural, fluent, and human-like in Vietnamese. Avoid rigid, repetitive, or machine-like expressions. \item Use lexical variation when appropriate. Prefer specific and natural Vietnamese expressions over overly generic wording. \item If the image clearly contains recognizable entities such as locations, brands, places, or objects, prioritize the specific entity name. Do not guess entity names when the visual evidence is unclear. \item Avoid uncertain or image-referential expressions such as ``có thể'', ``có lẽ'', ``dường như'', ``hình như'', ``dự đoán'', ``không rõ'', ``trong ảnh'', and ``bức ảnh cho thấy''. \end{enumerate} 
\promptsection{Output format} \begin{itemize}[leftmargin=*, itemsep=1pt, topsep=2pt] \item Hypothesis 1: ... \item Hypothesis 2: ... \end{itemize} \end{promptcard}

\subsection{Contradiction Generation Prompt} \label{app:contradiction_prompt} 

\begin{promptcard}{CONTRADICTION Prompt for Visual Entailment} 

\promptsection{Role} You are an expert in Vietnamese Visual Entailment data construction. Your task is to generate high-quality Vietnamese hypotheses for the Contradiction label. 

\promptsection{Label definition} A hypothesis is labeled as Contradiction if it conflicts with observable visual evidence in the image. 

\promptsection{Task} Given an image and entailment hypotheses derived from that image, generate two Vietnamese contradiction hypotheses. Each contradiction hypothesis should remain close to the original visual context but must contain one decisive semantic conflict with the image.

\promptsection{Input} \begin{itemize}[leftmargin=*, itemsep=1pt, topsep=2pt] \item Image: provided separately. \item Entailment hypothesis 1: provided separately. \item Entailment hypothesis 2: provided separately. \end{itemize} \promptsection{Generation procedure} \begin{enumerate}[leftmargin=*, itemsep=2pt, topsep=2pt] \item Observe the visible entities, quantities, colors, attributes, actions, and events in the image. \item Compare the entailment hypotheses with the image. \item Create each contradiction hypothesis by modifying exactly one key visual detail so that the new sentence becomes visually false. \item Keep the sentence natural and fluent in Vietnamese while preserving the general scene context. \end{enumerate} \promptsection{Allowed contradiction rules} Each generated hypothesis should use one of the following rules: \begin{enumerate}[leftmargin=*, itemsep=2pt, topsep=2pt] \item Replace an attribute with an opposite or incompatible attribute. \item Change the quantity of visible objects or people. \item Modify the time reference when the image provides a clear conflicting cue. \item Make an incorrect claim about a visible object. \item Make an incorrect claim about a visible event or action. \end{enumerate} \promptsection{Hard constraints} \begin{enumerate}[leftmargin=*, itemsep=2pt, topsep=2pt] \item The contradiction must be refutable from the image. Do not create a sentence that is merely unverifiable. \item Each contradiction hypothesis should contain one central semantic conflict. Avoid changing too many details at once. \item Do not rely on direct negation alone. Avoid superficial forms such as ``không'', ``chẳng'', ``chưa'', ``thay vì'', and ``nhưng'' as the main source of contradiction. \item Do not repeat the entailment sentence with only a trivial word change. The final sentence should be natural and independently readable. \item Avoid uncertain or image-referential expressions such as ``có thể'', ``có lẽ'', ``dường như'', ``hình như'', ``trong ảnh'', and ``bức hình''. \end{enumerate} \promptsection{Output format} \begin{itemize}[leftmargin=*, itemsep=1pt, topsep=2pt] \item Contradiction 1: ... \item Contradiction 2: ... \end{itemize} \end{promptcard} 

\subsection{Neutral Generation Prompt} \label{app:neutral_prompt} 
\begin{promptcard}{NEUTRAL Prompt for Visual Entailment} 

\promptsection{Role} You are an expert in Vietnamese Visual Entailment data construction. Your task is to generate high-quality Vietnamese hypotheses for the Neutral label. 

\promptsection{Label definition} A hypothesis is labeled as Neutral if it is plausible in the image context but cannot be confirmed or refuted based solely on the visual evidence. 

\promptsection{Task} Given an image and entailment hypotheses derived from that image, generate two Vietnamese neutral hypotheses. Each neutral hypothesis should introduce a specific new detail that is not visually verifiable from the image.

\promptsection{Input} \begin{itemize}[leftmargin=*, itemsep=1pt, topsep=2pt] \item Image: provided separately. \item Entailment hypothesis: provided separately. \end{itemize} \promptsection{Generation requirements} \begin{enumerate}[leftmargin=*, itemsep=2pt, topsep=2pt] \item The hypothesis must not be directly supported by the image. \item The hypothesis must not contradict visible evidence in the image. \item The hypothesis should remain plausible given the image context. \item Each neutral hypothesis should add exactly one unverifiable detail. \item The two neutral hypotheses should be generated using different neutral rules. \end{enumerate} \promptsection{Allowed neutral rules} Each generated hypothesis should use one of the following rules: \begin{enumerate}[leftmargin=*, itemsep=2pt, topsep=2pt] \item Add a specific time reference. \item Add a specific age or gender reference when it cannot be visually verified with certainty. \item Add a specific emotion, intention, or internal mental state. \item Add a specific size, distance, weight, area, or measurement. \item Add a related past or future event. \item State the purpose or intention behind a visible action. \end{enumerate} \promptsection{Rule-specific constraints} \begin{enumerate}[leftmargin=*, itemsep=2pt, topsep=2pt] \item If age is added, use a specific age expression such as ``35 tuổi'' or ``5 tuổi''. \item If measurement is added, include a clear measurement unit such as cm, m, kg, g, km, or lít. \item If time is added, use a specific time expression such as an hour, date, season, or year. \item If a past or future event is added, include a clear temporal cue such as ``hôm qua'', ``sắp tới'', ``tuần sau'', or ``ngày mai''. \item If a purpose or intention is added, use expressions such as ``muốn'', ``dự định'', ``hy vọng'', ``đang cố'', or ``nhằm''. \end{enumerate} \promptsection{Hard constraints} \begin{enumerate}[leftmargin=*, itemsep=2pt, topsep=2pt] \item Do not use the same neutral rule for both generated hypotheses. \item Decide the neutral rule before writing the sentence. \item Use natural Vietnamese and avoid unnatural or overly generic phrasing. \item Avoid uncertain or image-referential expressions such as ``có thể'', ``có vẻ'', ``có lẽ'', ``dường như'', and ``có khả năng''. \end{enumerate} \promptsection{Output format} \begin{itemize}[leftmargin=*, itemsep=1pt, topsep=2pt] \item Neutral 1: ... \item Neutral 2: ... \end{itemize} \end{promptcard}

\end{document}